\documentclass[iicol,sn-apa]{sn-jnl}

\usepackage{multirow}%
\usepackage{amsmath,amssymb,amsfonts}%
\usepackage{amsthm}%
\usepackage{graphicx}
\usepackage{mathrsfs}%
\usepackage[title]{appendix}%
\usepackage{xcolor}%
\usepackage{subfigure}
\usepackage{textcomp}%
\usepackage{manyfoot}%
\usepackage{booktabs}%
\usepackage{adjustbox}
\usepackage{algorithm}%
\usepackage{algorithmicx}%
\usepackage{algpseudocode}%
\usepackage{listings}%
\usepackage{xspace}
\newcommand{\framework}{SEPL\xspace}

\theoremstyle{thmstyleone}%
\newcommand{\algorithmname}{Algorithm}
\theoremstyle{thmstyletwo}%

\theoremstyle{thmstylethree}%

\usepackage{todonotes}
\usepackage{xcolor}

\begin{document}

\title[Article Title]{Self-Ensemble Post Learning for Noisy Domain Generalization}


\author[1]{\fnm{Wang} \sur{Lu}}\email{newlw230630@gmail.com}

\author*[2]{\fnm{Jindong} \sur{Wang}}\email{jwang80@wm.edu}

\affil[1]{\orgname{Independent}, \orgaddress{\street{Haidian District}, \city{Beijing}, \postcode{100190}, \country{China}}}

\affil[2]{\orgname{William \& Mary}, \orgaddress{\street{} \city{Williamsburg}, \postcode{23185}, \state{VA}, \country{USA}}}


\abstract{
While computer vision and machine learning have made great progress, their robustness is still challenged by two key issues: data distribution shift and label noise. 
When domain generalization (DG) encounters noise, noisy labels further exacerbate the emergence of spurious features in deep layers, i.e. spurious feature enlargement, leading to a degradation in the performance of existing algorithms.
This paper, starting from domain generalization, explores how to make existing methods rework when meeting noise. 
We find that the latent features inside the model have certain discriminative capabilities, and different latent features focus on different parts of the image. 
Based on these observations, we propose the \textbf{S}elf-\textbf{E}nsemble \textbf{P}ost \textbf{L}earning approach (\framework) to diversify features which can be leveraged. 
Specifically, \framework consists of two parts: feature probing training and prediction ensemble inference.  
It leverages intermediate feature representations within the model architecture, training multiple probing classifiers to fully exploit the capabilities of pre-trained models, while the final predictions are obtained through the integration of outputs from these diverse classification heads. 
Considering the presence of noisy labels, we employ semi-supervised algorithms to train probing classifiers. 
Given that different probing classifiers focus on different areas, we integrate their predictions using a crowdsourcing inference approach. 
Extensive experimental evaluations demonstrate that the proposed method not only enhances the robustness of existing methods but also exhibits significant potential for real-world applications with high flexibility.
}





\maketitle


\section{Introduction}\label{sec-intro}

In recent years, deep learning has made significant progress and has been widely applied in various fields, particularly in healthcare domains~\citep{delussu2024synthetic,jianggraphcare}. 
However, a successful deep learning model often has stringent requirements for the scenario and data, such as data being from the same distribution and being free of noise. 
These conditions are often challenging to meet in real-world applications, especially in healthcare~\citep{baek2024unexplored,zheng2024exploiting}.

\begin{figure*}[!t]
\centering
\includegraphics[width=0.8\textwidth]{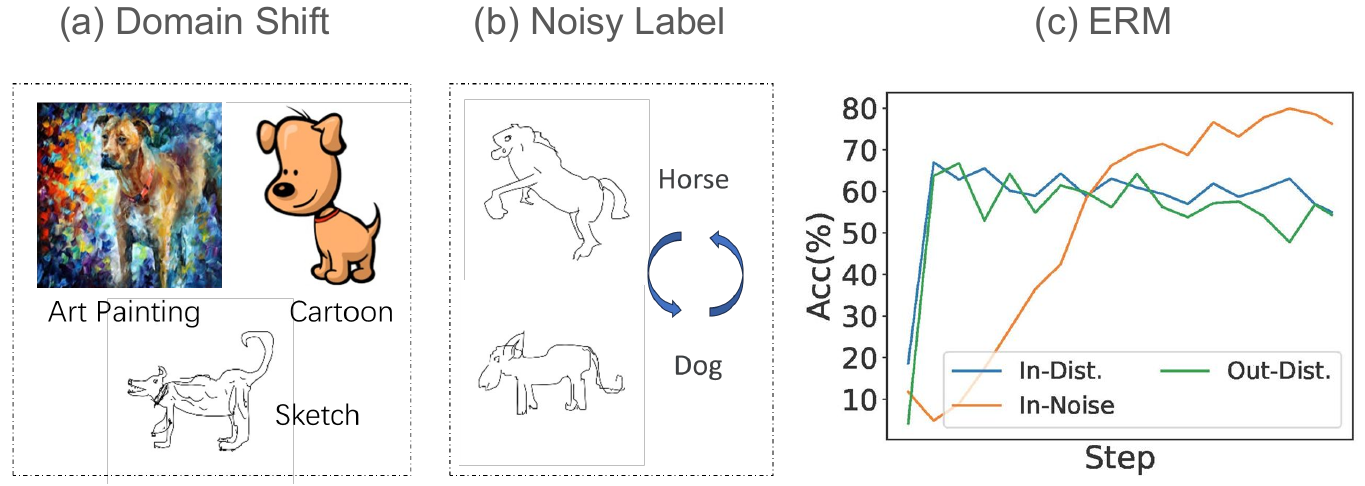}
\caption{Illustration of data distribution shift and label noise:
(a) Three images showing pictures of different dogs with varying styles.
(b) People might confuse the sketches of dogs and horses.
(c) The process of training a model directly without addressing distribution shift and noisy labels. (Following \citep{qiaounderstanding}, we flip $25\%$ of training data labels.)}\label{fig-main-intro-fig1}
\end{figure*}

In real-world medical environments, due to various factors such as differences in perspectives, machines, and individual patient variations, the collected pathological data often exhibit distribution discrepancies~\citep{medmnistv2,bilic2023liver}. 
Additionally, even professional clinicians cannot achieve $100\%$ accuracy in labeling diseases~\citep{graber2005diagnostic}. 
Therefore, this paper attempts to focus on these two common issues encountered in practice. 
To further understand these two issues, we provide examples on PACS~\citep{li2017deeper} in \figureautorefname~\ref{fig-main-intro-fig1}(a)-(b).
As shown in \figureautorefname~\ref{fig-main-intro-fig1}(a), the three images have different styles, which can be interpreted as shifts in the input feature distributions. 
While \figureautorefname~\ref{fig-main-intro-fig1}(b) presents that people might confuse the sketches of dogs and horses.
If we ignore these two issues and train the model directly, it may lead to poor results. 
As seen in \figureautorefname~\ref{fig-main-intro-fig1}(c), the training accuracy within the same distribution remains unchanged, while the test accuracy outside the distribution gradually decreases. 
Additionally, the model increasingly fits to the noisy labels. 
This severely hinders the practical application of deep learning in real-world settings~\citep{song2022learning,zhou2022domain}.

In the past years, distribution shift and noisy label learning are two popular, but rather \emph{isolated} research areas. 
On the one hand, researchers typically refer to the distribution shift as the out-of-distribution (OOD) problem that is often addressed using domain generalization (DG) methods~\citep{wang2022generalizing}. 
Existing DG methods can be broadly classified into three types: data manipulation, representation learning, and learning strategy. 
For example, the classical method CORAL learns domain-invariant representations by aligning the covariance matrices of the representations~\citep{sun2016deep}, while Model Ratatouille enhances generalization ability through model combination~\citep{rame2023model}. 
These methods have shown some capability in classic domain generalization tasks, but they do not take label noise into account.
On the other hand, learning with noisy labels is also a popular research area~\citep{song2022learning,bucarelli2023leveraging}. 
Common solutions typically involve identifying data that is likely to be noisy, treating them as unlabeled, and then solving the problem using semi-supervised methods. 
For example, OT-Filter~\citep{feng2023ot} uses optimal transport algorithms within the EM framework to detect noise and applies MixMatch~\citep{berthelot2019mixmatch} for semi-supervised learning. 

\begin{figure*}[!t]
\centering
\includegraphics[width=0.7\textwidth]{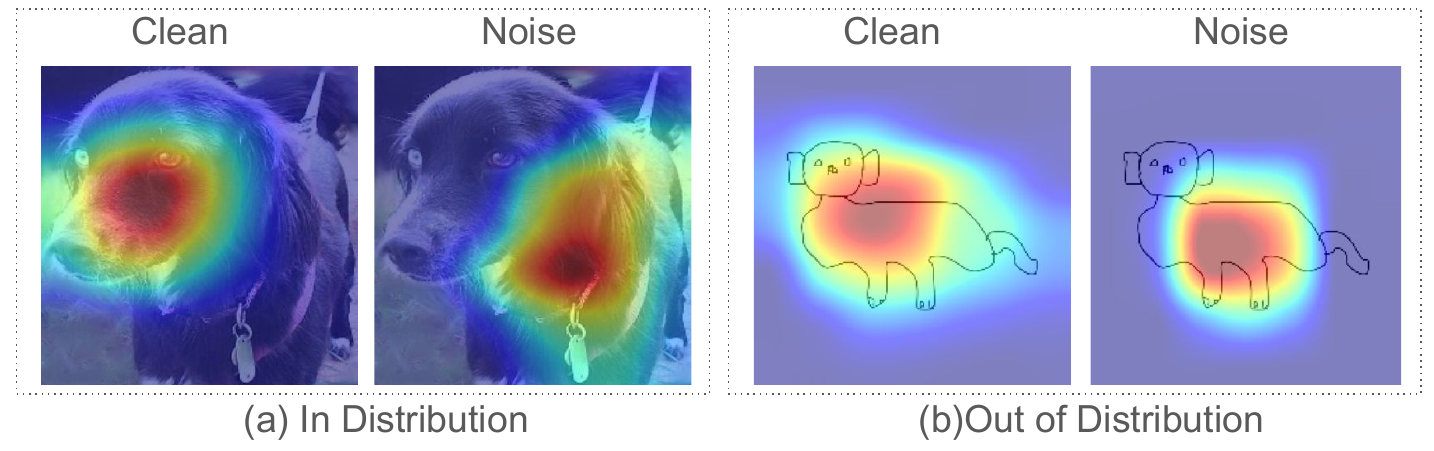}
\caption{Grad-CAM diagram under the influence of data distribution shift and noisy labels}
\label{fig-main-intro-fig2}
\end{figure*}

Although this field has seen significant progress, there are few studies that simultaneously address both noise and distribution shift, despite the fact that these issues are often common in real-world applications. 
Recently, \citep{qiaounderstanding} attempted to analyze out-of-distribution (OOD) scenarios through noise analysis, finding that noise degrades the performance of OOD methods. And \citep{sanyal2024accuracy} even discovered that noisy labels further increase the difficulty of extracting meaningful features. 
As shown in \figureautorefname~\ref{fig-main-intro-fig2}, compared to models trained on clean data, models trained on noisy data tend to focus more on non-critical areas, such as the dog's body. 
We refer to this phenomenon as spurious feature entanglement.
As far as we know, no attempt has been made to make existing OOD methods work again.

To make existing OOD methods rework when meeting noisy labels, we endeavor to leverage more diversify and effective features.
We first carefully analyze the model's internal feature capabilities when both distribution shift and noise are present. 
We find that the model's internal features have some discriminatory power, and the focal areas of the features at different layers vary. 
Based on these observations, we propose a generic self-ensemble post learning method, \framework, that does not require retraining the model backbone from scratch.
\framework fully leverages the backbone's capabilities to perform ensemble decisions, offering efficiency and lower requirements for training hardware. 
Considering the presence of noise in the labels, we introduce a semi-supervised approach by treating the model's low-confidence predictions as unlabeled data. 
Taking into account that different layers of the model focus on different image features, we incorporate a crowdsourcing-based ensemble method~\citep{dawid1979maximum}.
Extensive experiments and real-world applications demonstrate the strong performance of our method under distribution shift and label noise scenarios. Our contributions are as follows:
\begin{itemize}
    \item \textbf{Insightful observations:} We found that the model's internal features have some discriminatory power, and there are differences in the parts of the image that each feature focuses on.
    \item \textbf{High-performing method:} Based on observations, we propose an efficient post-processing model self-ensemble method, \framework, to cope with noisy domain generalization. \framework consists of two parts: feature probing training and prediction ensemble inference. It is highly flexible and can be adapted to integrate different algorithms based on specific requirements.
    \item \textbf{Superior performance and insightful results:} Comprehensive experimental results showcase the advantages of our method and its promising potential for real-world applications.
\end{itemize}

The remainder of this paper is organized as follows.
We will review related work in \sectionautorefname~\ref{sec-related} while we will present our methodology in \sectionautorefname~\ref{sec-method}. 
\sectionautorefname~\ref{sec-exp} will provide experimental comparisons, followed by real-world application results in \sectionautorefname~\ref{sec-app}. 
In \sectionautorefname~\ref{sec-limit}, we will discuss the limitations of existing methods. 
Finally, \sectionautorefname~\ref{sec-conclu} will conclude the paper.

\section{Related Work}\label{sec-related}

\subsection{Domain generalization}

Solutions to data distribution shift are typically approached using transfer learning methods~\citep{pan2009survey}. 
Transfer learning can be divided into domain adaptation and domain generalization, depending on whether the target data is available during training~\citep{wang2018deep,wang2022generalizing}. 
In real-world scenarios, the target data is often unavailable, which is why this paper focuses on the domain generalization (DG) setting~\citep{liu2023ss}.

As discussed in the literature, the goal of domain generalization is to train a model on source data that can perform well on unseen, target data that may have different distributions. 
Existing methods for domain generalization can be categorized into three types: data manipulation, representation learning, and learning strategies.
Data manipulation primarily involves augmenting or directly generating input data. 
For example, the classical Mixup algorithm generates new data by performing linear interpolation between two instances and their labels, with the interpolation weight sampled from a Beta distribution~\citep{zhang2018mixup}. 
This method does not require training generative models. 
Later, methods like LISA improved upon Mixup by restricting sample selection and generation techniques to better create samples~\citep{yao2022improving}. 
In recent years, some approaches have used large models to assist in generating more diverse data, thereby enhancing model generalization~\citep{li2024beyond}. 
By incorporating large language models' understanding of categories, domains, environments, and other contextual information, the generated data tends to be more realistic and diverse.
Representation learning can be further divided into domain-invariant feature learning and feature disentanglement. 
For example, ADRMX addresses the issue of ignoring domain-specific features by incorporating both domain-variant and domain-invariant features using an original additive disentanglement strategy~\citep{demirel2023adrmx}. 
StableNet learnt weights for training samples to remove the dependencies between features, which helped deep models get rid of spurious correlations and, in turn, concentrated more on the true connection between discriminative features and labels~\citep{zhang2021deep}. 
ManyDG treated each patient as a separate domain, identified the patient domain covariates by mutual reconstruction, and removed them via an orthogonal projection step~\citep{yangmanydg}.
More recently, some approaches have tried to integrate multimodal information into representations, such as CLIPCEIL~\citep{yuclipceil}.
Learning strategies primarily focus on specialized techniques such as meta-learning, ensemble learning, and gradient manipulation. 
Fishr enforcesd domain invariance in the space of the gradients of the loss and eventually aligned the domain-level loss landscapes locally around the final weights~\citep{rame2022fishr}. 
DiWA averaged weights obtained from several independent training runs with differences in hyperparameters and training procedures~\citep{rame2022diverse}.
Model Ratatouille repurposed the auxiliary weights as initializations for multiple parallel fine-tunings on the target task, and then averaged all fine-tuned weights to obtain the final model~\citep{rame2023model}.

Although domain generalization has made significant progress, few methods have focused on scenarios where label noise is present, which remains a significant gap in the field.

\begin{figure*}[!t]
\centering
\includegraphics[width=.8\textwidth]{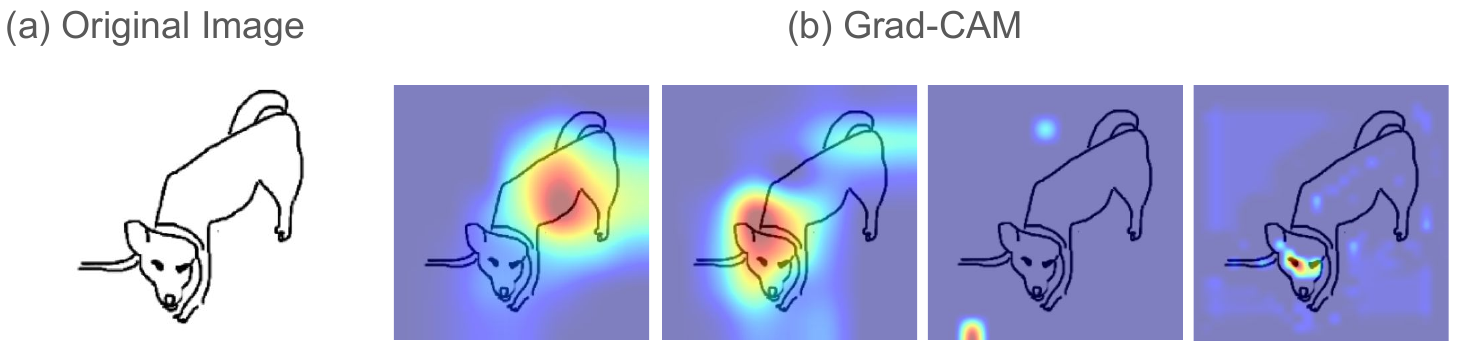}
\caption{Visualization of attention across different layers. (a) Test input image. (b) Grad-CAM attention maps derived from features of different layers using linear probes.}
\label{fig-main-mot1}
\end{figure*}

\subsection{Noisy label learning}

Label noise is a classic topic that can be addressed by various approaches, such as improved network architectures, regularization techniques, loss function design, loss adjustments, and sample selection~\citep{song2022learning,yisource}. 
SDM exploited the self-cognition ability of neural networks to denoise during training and built a selective distillation module to optimize computational efficiency~\citep{sunself}.
LSL incorporated additional distribution information—structural labels with a revers k-NN to help the model in achieving a better feature manifold and mitigating overfitting to noisy labels~\citep{kim2024learning}.
PLM encouraged the model to focus on and integrate richer information from various parts~\citep{zhao2024estimating}.
It partitioned features into distinct parts by cropping instances, and introduced a single-to-multiple transition matrix to model the relationship between the noisy and part-level labels.
OT-Filter provided geometrically meaningful distances and preserved distribution patterns to measure the data discrepancy, which revamped the sample selection~\citep{feng2023ot}.
While CSOT considered the inter- and intra-distribution structure of the samples to construct a robust denoising and relabeling allocator based on curriculum and structure-aware optimal transport~\citep{chang2023csot}.
Moreover, noisy label learning is often combined with other scenarios, such as in federated learning.
FedDiv proposed a global noise filter for effectively identifying samples with noisy labels on every client and introduced a Predictive Consistency based Sampler to identify more credible local data for model training~\citep{li2024feddiv}.

\subsection{Noisy domain generalization}
Some approaches have attempted to analyze label noise in conjunction with out-of-distribution (OOD) scenarios. 
\citep{humblot2024noisy} taked a closer look at OOD detection methods in the scenario where the labels used to train the underlying classifier were unreliable.
\citep{sanyal2024accuracy} found that noisy data and nuisance features could disrupt the positively correlated relationship between in-distribution (ID) and out-of-distribution (OOD) accuracy, leading to a negative correlation, termed Accuracy-on-the-wrong-line.
\citep{qiaounderstanding} investigated whether there were benefits of DG algorithms over ERM through the lens of label noise.
\citep{ji2023drugood} presented a systematic OOD dataset curator and benchmark for AI-aided drug discovery with distribution shifts and noise existences.
And \citep{ma2024sharpness} proposed SAGA for domain generalization with
noisy labels in fault diagnosis using  dual network structure.
Moreover, \citep{albert2022embedding} proposed a two-stage algorithm leveraging contrastive feature learning and clustering to separate OOD and ID noisy samples, improving robustness in training on web-crawled image datasets, while \citep{albert2022addressing} analyzed web label noise distribution and introduced Dynamic Softening of OOD Samples (DSOS) to bridge the gap between noisy and fully clean datasets.
There is no existing method that specifically considers how to enhance existing domain generalization (DG) approaches, enabling them to function effectively in noisy domain generalization and achieve practical applicability in real-world scenarios.

\begin{figure*}[t!]
    \centering
    \subfigure[Training Acc. (ID)]{
        \includegraphics[height=0.16\textwidth]{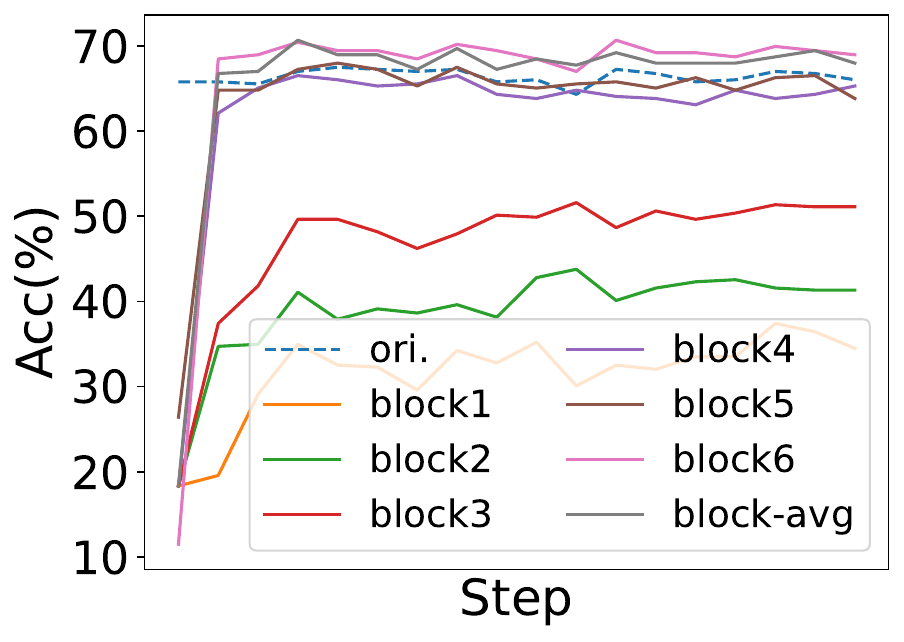}
        \label{fig-mot2-1}
    }
    \subfigure[Noisy Acc. (ID)]{
        \includegraphics[height=0.16\textwidth]{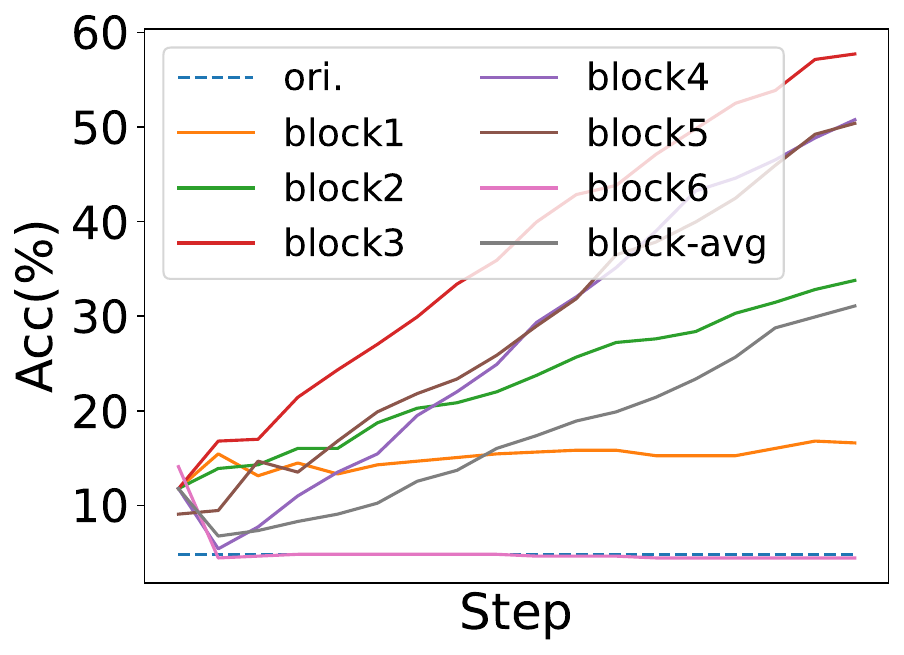}
        \label{fig-mot2-2}
    }
    \subfigure[Test ACC. (OOD)]{
        \includegraphics[height=0.16\textwidth]{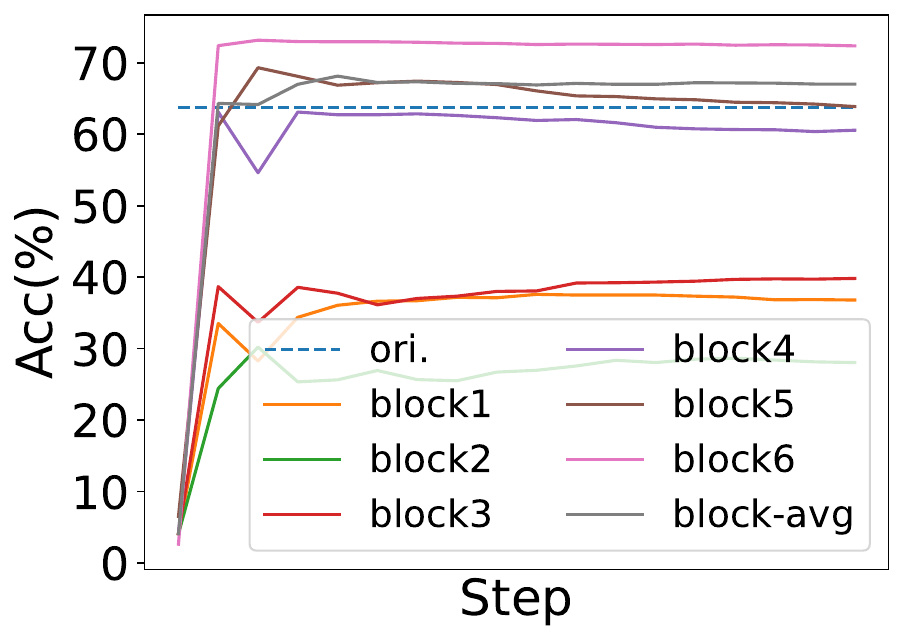}
        \label{fig-mot2-3}
    }
    \subfigure[SL VS SSL]{
        \includegraphics[height=0.16\textwidth]{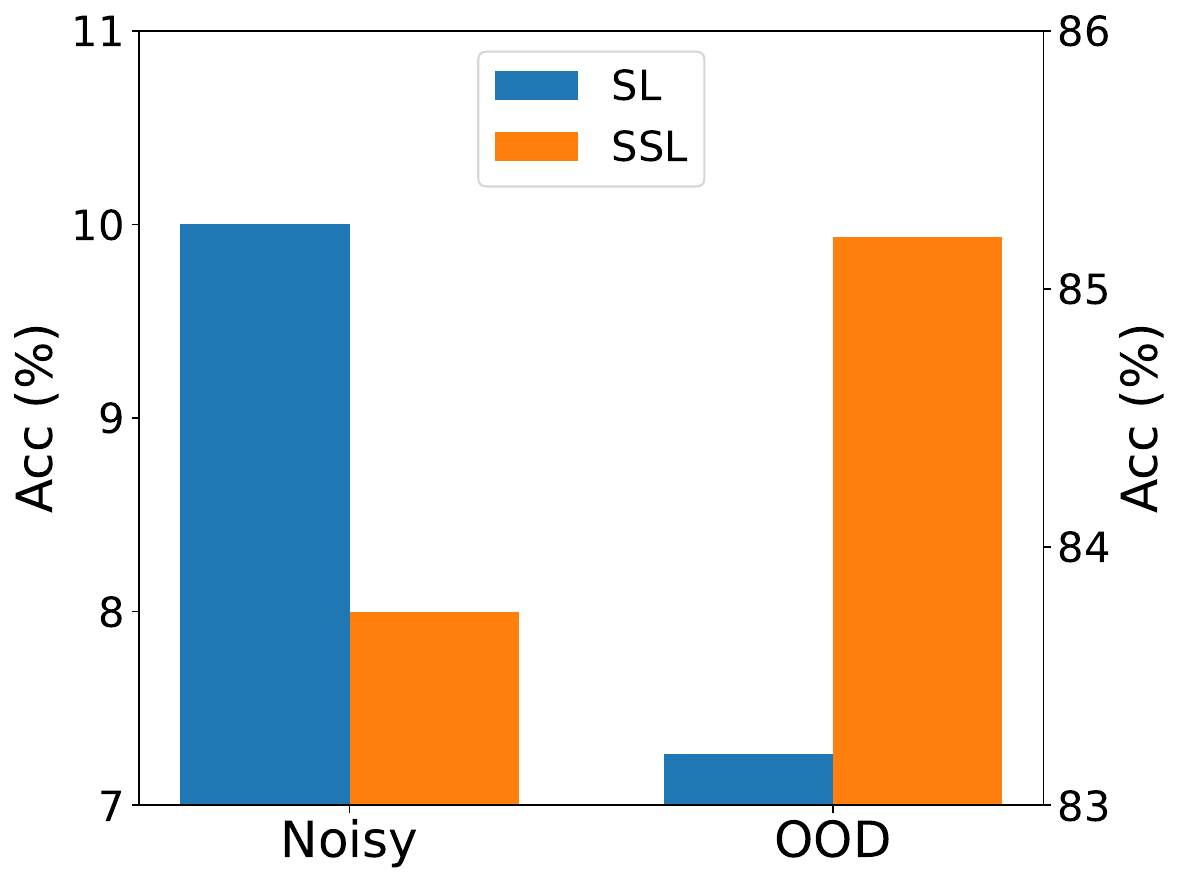}
        \label{fig-mot2-4}
    }
    \caption{Accuracy during the training process. (a) The accuracy variation on in-distribution data. (b) The accuracy variation on noisy data. (c) The accuracy variation on out-of-distribution data. (d) The accuracy on OOD and Noisy ID data using supervised learning (SL) and semisupervised learning (SSL).}
    \label{fig-mot2}
\end{figure*}

\section{Methodology}\label{sec-method}
\subsection{Problem Formulation}
We adopt the problem formulation for domain generalization as outlined in \citep{wang2022generalizing}, focusing on a $C$-class classification scenario. 
In this scenario, we are provided with multiple labeled source domains, denoted as $\mathcal{S} = \{ \mathcal{S}^i \mid i = 1, \dots, M \}$, where $M$ represents the total number of source domains. Each source domain $\mathcal{S}^i = \{(\mathbf{x}_j^i, y_j^i)\}_{j=1}^{n_i}$ corresponds to the $i^{th}$ domain, with $n_i$ indicating the number of samples in $\mathcal{S}^i$. 
Notably, the joint distributions $P^i_{XY}$ and $P^j_{XY}$ differ across domains, i.e., $P^i_{XY} \neq P^j_{XY}$ for $1 \leq i \neq j \leq M$. 
The objective of domain generalization is to learn a predictive function $h: \mathcal{X} \rightarrow \mathcal{Y}$ that generalizes well across the $M$ source domains, aiming to minimize the prediction error on an unseen target domain $\mathcal{S}_{test}$, whose joint distribution is unknown. 
This is formally expressed as $\min_{h} \mathbb{E}_{(\mathbf{x},y)\in \mathcal{S}_{test}} [\ell(h(\mathbf{x}),y)]$, where $\mathbb{E}$ denotes the expectation and $\ell(\cdot, \cdot)$ is the loss function. 
All domains, including both source and target domains, share the same input and output spaces: $\mathcal{X}^1 = \cdots = \mathcal{X}^M = \mathcal{X}^T \in \mathbb{R}^{m}$, where $\mathcal{X}$ is the $m$-dimensional instance space, and $\mathcal{Y}^1 = \cdots = \mathcal{Y}^M = \mathcal{Y}^T = \{1, 2, \dots, C\}$, where $\mathcal{Y}$ represents the label space.
Please note that the only difference is the presence of label noise in the training data, meaning there is a possibility of mislabeling in $y$.

\subsection{Motivation}

Taking ResNet50~\citep{he2016deep} as an example, we analyze our motivation on PACS.
Following \citep{qiaounderstanding}, we flipped $25\%$ of the training data labels.
We conduct linear probing on the input to layer1, the output of layer1, layer2, layer3, layer4, and the final pooled output features to examine the characteristics of intermediate features.

\paragraph{Are the focal areas of features across different layers the same?}
From \figureautorefname~\ref{fig-main-mot1}, it is evident that features detected by different layers focus on varying aspects. 
For instance, in \figureautorefname~\ref{fig-main-mot1}(a), the original model primarily focuses on the body of the dog, while the final output mainly attends to the dog's head and tail. 
Meanwhile, intermediate outputs may focus on elements like eyes or the surrounding environment. 
This indicates that features from different layers have distinct attention points.

\paragraph{Do the features from different layers have discriminative capability?}
From \figureautorefname~\ref{fig-mot2}(a)-(c), we can observe that features from different layers indeed exhibit discriminative power. 
\figureautorefname~\ref{fig-mot2}(a) shows that as the layer depth increases, the classification performance of the features improves. 
The first three layers generally have weaker discriminative ability, while the last few layers are more discriminative. 
\figureautorefname~\ref{fig-mot2}(b) illustrates that as the depth increases, the model are more likely to fit noise.
However, selecting an appropriate initial model offers an advantage for the deeper layers (as seen in \figureautorefname~\ref{fig-main-intro-fig1}(c), prolonged training of the initial model increases noise fitting; here, we select the initial model using validation data and apply early stopping).
\figureautorefname~\ref{fig-main-intro-fig2} and \figureautorefname~\ref{fig-main-mot1} show that as training progresses, the focal areas of deep-layer features on OOD data shifts, leading to redundant features and a decrease in accuracy (as shown in \figureautorefname~\ref{fig-main-intro-fig1}). 
Therefore, it is essential to identify generalizable and useful features by incorporating diverse perspectives.
\figureautorefname~\ref{fig-mot2}(c) demonstrates that combining outputs from different layer probes can improve out-of-distribution generalization.
\footnote{On one hand, different layers exhibit certain discriminative abilities on OOD data; on the other hand, simple averaging even outperforms the initial results.}

\paragraph{Does label noise affect the training of intermediate feature probing classifiers?} 
Even though shallow features exhibit a certain degree of robustness to noise, they are still affected by noise during probing.
As shown in \figureautorefname~\ref{fig-mot2}(d), even when selecting a well-performing pre-trained model based on validation performance, the ensemble inference results of intermediate feature probing classifiers are still influenced by noisy labels. 
When training directly with noisy labels, the noise fitting accuracy is $10\%$, and the OOD accuracy is $83.2\%$. 
However, if the dataset is divided into clean labeled data and unlabeled noisy data for semi-supervised training, the noise fitting accuracy drops to $8\%$, while the OOD accuracy increases to $85.2\%$. 
This indicates that label noise interferes with probing results, and proper data processing is necessary.

These observations inspire us to leverage the outputs of intermediate feature layers within the model to enhance robustness against label noise while simultaneously improving generalization performance.
Moreover, during training probing classifiers, semi-supervised learning is required to mitigate the impact of noisy data.

\begin{figure*}[!t]
\centering
\includegraphics[width=\textwidth]{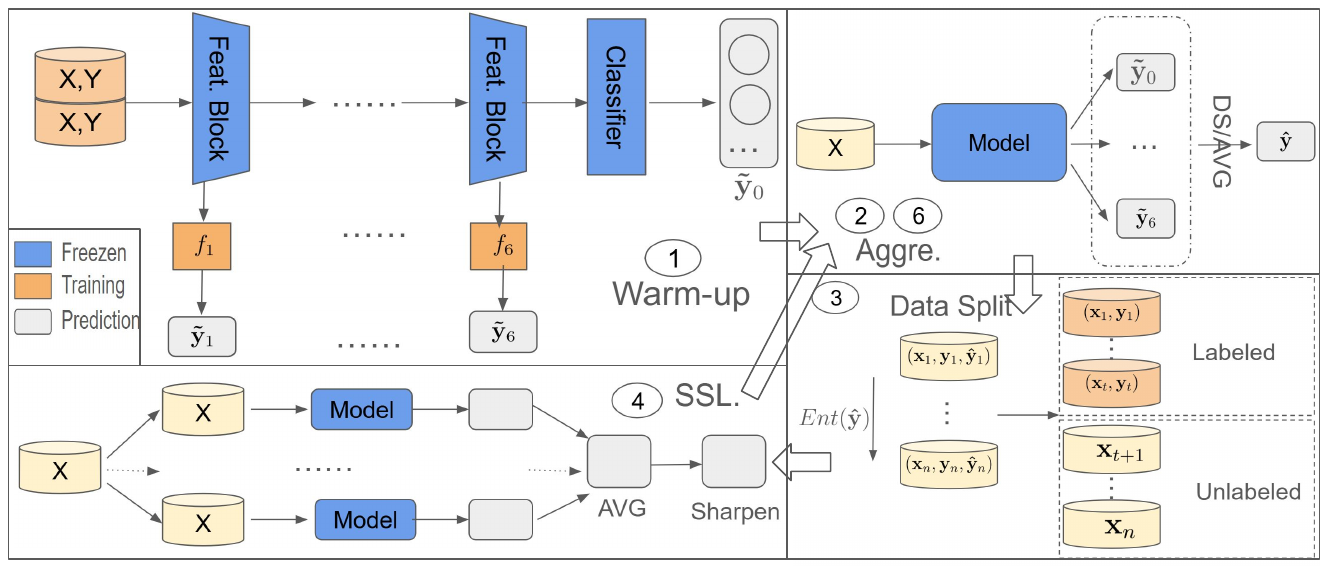}
\caption{The diagram of \framework.}\label{fig-main-framework}
\end{figure*}

\subsection{Our Approach}
In this section, we introduce the proposed method (using ResNet50 as an example).

Based on the analysis above, we know that the intermediate hidden features of a model possess certain discriminative abilities. 
Meanwhile, different layers focus on distinct aspects of images, specifically, shallow layers exhibit greater robustness to noise.
Therefore, our primary goal is to fully utilize the intermediate features of a pretrained model and perform probing on these features.
Motivated by this, we designed a post-processing method based on model self-ensemble. 
The proposed method conducts independent classification on the intermediate features of a pre-trained model and consists of feature probing training and prediction ensemble inference.

\paragraph{Feature probing training}
During training, the pre-trained model, $h=(g \circ f)$, is frozen and used solely for feature extraction. 
To ensure diverse feature representation, we divide the feature module of the pre-trained model into $T$ parts ($T=6$ for ResNet-50), i.e. $f= f_T\circ f_{T-1} \circ \ldots  \circ f_1$. 
After extracting features from each part, we perform probing by training a simple probing head to map the features to the target task.
For the $i^{th}$ feature $\mathbf{z}_i$, we have
\begin{equation}
    \mathbf{\tilde{y}}_i=g_i(\mathbf{z}_i) = g_i(f_i(\mathbf{x}_i)),
\end{equation}
where $g_i$ and $\mathbf{z}_i$ denote the $i^{th}$ probing classifier and extracted features, respectively, and $\mathbf{\tilde{y}}_i$ denotes the prediction of the probing classifier.\footnote{
Unless otherwise specified, $\mathbf{\tilde{y}}_i$ refers to a soft label, meaning it is an encoded vector.} 
Similar to common training for classification, we utilize the crossentropy loss to optimize $g_i$.
In this way, we obtain more robust and diverse predictions from the intermediate layer features, e.g. $(\mathbf{x},\mathbf{\tilde{y}}_0,\mathbf{\tilde{y}}_1,\cdots,\mathbf{\tilde{y}}_T)$. 
Here, $\mathbf{\tilde{y}}_0$ denotes the original prediction of the pre-trained model.
This method can be implemented with straightforward direct training besides iteratively combined with different methods.

Considering the presence of noisy labels, directly training with noisy labels would result in poor performance of the probe classifiers. 
Therefore, we follow common noise learning frameworks to first identify noisy samples and then use semi-supervised methods for learning. 
Referring to \citep{feng2023ot}, we utilize MixMatch for robust semi-supervised training, but with fewer operations. 
For clean samples, their labels are retained, while noisy sample labels are disregarded. 
The training process then employs both labeled and unlabeled data in a semi-supervised learning framework.
Please note that we do not update the feature extraction part of the model or the original classifier; we only update the intermediate feature probing classifiers.

\paragraph{Prediction ensemble inference}
During inference, we collect the classification outputs from each probing head, $\mathbf{\tilde{y}}_i$, and perform ensemble operations.
And we have the final prediction as the following,
\begin{equation}
    \mathbf{\hat{y}}=\mathcal{AGG}(\mathbf{\tilde{y}}_0,\mathbf{\tilde{y}}_1,\mathbf{\tilde{y}}_2,\cdots,\mathbf{\tilde{y}}_T).
\end{equation}
The ensemble operator $\mathcal{AGG}$ can be as simple as averaging, i.e., $ \mathbf{\hat{y}}=\frac{1}{T+1}\sum_{i=0}^{T} \mathbf{\tilde{y}}_i,$
or it can leverage more sophisticated crowdsourcing approaches. 

In \framework, given the different abilities and focal area of the probing classifiers mentioned in the motivation section, we employ the Dawid-Skene (DS)~\citep{dawid1979maximum} algorithm.
DS is a probabilistic model that iteratively estimates the true labels of data points and the error rates of individual annotators by initializing label probabilities, updating annotator confusion matrices based on current label estimates, and then refining the label probabilities using the weighted contributions of annotators until convergence.

\paragraph{Process flow}
Based on the two parts mentioned above, \figureautorefname~\ref{fig-main-framework} illustrates the overall workflow of our method: 
\footnote{Step 2-5 correspond to feature probing training while Step 6 corresponds to prediction ensemble inference. In addtion, in Step 4, we only present the labeling process for the unlabeled data, which is largely similar to the original MixMatch method~\citep{berthelot2019mixmatch}.}
\begin{enumerate}
    \item \textbf{Warm-up}: Load the training data and the pre-trained model. Train the probing classifiers using the training data.
    \item \textbf{Training aggregation}: Aggregate the predictions of probing classifiers along with the original predictions from the pre-trained model. 
    \item \textbf{Dataset splitting}: Use the aggregation to split the original dataset into two subsets: noise-free data (labeled) and potentially noisy data (unlabeled).
    \item \textbf{Semi-supervised training}: Use the MixMatch semi-supervised learning algorithm to update the probing classifiers.
    \item \textbf{Iteration}: Repeat steps 2-4 until convergence or reaching the maximum rounds.
    \item \textbf{Inference aggregation}: After obtaining the predicted labels from all probing classifiers, we perform ensemble inference on the prediction results. This can be done by directly averaging the predictions or using a crowdsourcing inference method.
\end{enumerate}
Next, we will provide some detailed explanations.

\paragraph{Aggregation}

Assume we have labeled data $\mathcal{D} = \{(\mathbf{x},\mathbf{\tilde{y}}_0,\mathbf{\tilde{y}}_0,\cdots,\mathbf{\tilde{y}}_6)\}_{i=1}^N$, where $\mathbf{\tilde{y}}_j$ is the prediction of the $j^{th}$ probing classifiers and $N$ is the number of samples. 
For the purpose of generalization, we assume $K=7$, where $K$ represents the number of annotators.
Since the labels are derived from probes at different hidden layers, we denote the classification ability of the probe at the $i^{th}$ hidden layer as $\boldsymbol{\pi}_i \in \mathbb{R}^{C\times C}$. 
Here, $\boldsymbol{\pi}_i(c,c')$ represents the probability that the probing classifier for the $i^{th}$ hidden feature classifies a sample with the true label $c$ as $c'$.
Our task now is to estimate the true labels, $argmax_c \mathbb{P}(\mathbf{\hat{y}}_i=c)$, of the samples based on the observed classifications, $\mathcal{D}$.
Thus, we have the following claim.
For the sample $\mathbf{x}_i$, the likelihood of the $\mathbf{\tilde{y}}_{i,j}$ can be,
\begin{equation}
\begin{aligned}
&\mathbb{P}(\mathbf{\tilde{y}}_{i,j}|\mathbf{\hat{y}}_i,\boldsymbol{\pi}_j)    \\
=& \Pi_{c'=1}^C(\sum_{c=1}^C \mathbf{\hat{y}}_{i,c}\boldsymbol{\pi}_j(c,c'))^{\mathbf{\tilde{y}}_{i,j,c'}}
\end{aligned}
\end{equation}
For all annotations of the $\mathbf{x}_i$, we have, 
\begin{equation}
    \begin{aligned}
&\mathbb{P}(\{\mathbf{\tilde{y}}_{i,j}\}|\mathbf{\hat{y}}_i,\{\boldsymbol{\pi}_j\})    \\
=&\Pi_{j=1}^K \Pi_{c'=1}^C(\sum_{c=1}^C \mathbf{\hat{y}}_{i,c}\boldsymbol{\pi}_j(c,c'))^{\mathbf{\tilde{y}}_{i,j,c'}}
    \end{aligned}
\end{equation}
Therefore, the final aim can be
\begin{equation}
\begin{aligned}
    &\max_{\{\mathbf{\hat{y}}_i\}, \{ \boldsymbol{\pi}_j\}}\mathcal{L}( \{\mathbf{\tilde{y}}_{i,j}\} | \{\mathbf{\hat{y}}_i\}, \{ \boldsymbol{\pi}_j\} )\\
    = &\Pi_{i=1}^N\Pi_{c=1}^C\mathbb{P}(\mathbf{\hat{y}}_i=c)^{\mathbf{\hat{y}}_{i,c}}
    \mathbb{P}(\{\mathbf{\tilde{y}}_{i,j}\}|\mathbf{\hat{y}}_i,\{\boldsymbol{\pi}_j\}).
\end{aligned}
\end{equation}

If not otherwise specified, we assume the prior distribution to be uniform. 
Note that, unlike the original DS, we generalize the concept of hard labels to soft labels, which allows us to better leverage probabilistic information. 
Next, we follow EM to solve the problem as in DS.

First, in the E-step, we estimate the true label distribution $\mathbf{\hat{y}}_i$.
Based on the current annotations $\mathbf{\tilde{y}}_{i,j}$ and the current estimation of $\boldsymbol{\pi}_j$, we update the true label distribution for sample $i$
\begin{equation}
\begin{aligned}
\mathbf{\hat{y}}_{i,c}  \propto \mathbb{P}(\mathbf{\hat{y}}_i=c)\Pi_{j=1}^K\Pi_{c'=1}^C
(\boldsymbol{\pi}_j(c,c'))^{\mathbf{\tilde{y}}_{i,j,c'}}.
\end{aligned}
\end{equation}
After normalization, we obtain the updated distribution,
\begin{equation}
\mathbf{\hat{y}}_{i,c} = \frac{ \mathbb{P}(\mathbf{\hat{y}}_i=c)\Pi_{j=1}^K\Pi_{c'=1}^C
(\boldsymbol{\pi}_j(c,c'))^{\mathbf{\tilde{y}}_{i,j,c'}}}
{\sum_{c'=1}^C\mathbb{P}(\mathbf{\hat{y}}_i=c')\Pi_{j=1}^K\Pi_{c''=1}^C
(\boldsymbol{\pi}_j(c',c''))^{\mathbf{\tilde{y}}_{i,j,c''}}}.
\label{eqa-main-updy-norm}
\end{equation}

In the M-step, we update the capability matrix $\boldsymbol{\pi}_j$. 
Using the annotations $\mathbf{\tilde{y}}_{i,j}$ and the estimated true labels $\hat{y}_{i,c}$, we have
\begin{equation}
\boldsymbol{\pi}_j(c,c')=\frac{\sum_{i=1}^N \mathbf{\hat{y}}_{i,c}\cdot \mathbf{\tilde{y}}_{i,j,c'}}{\sum_{i=1}^N\mathbf{\hat{y}}_{i,c}}.
\label{eqa-main-updpi}
\end{equation}

Here, we provide the algorithm of aggregation in \algorithmname~\ref{alg-main-ds}.
\footnote{This algorithm can be applied in both training and inference stages. 
If soft logits are used, we refer to it as SoftDS; otherwise, it is called HardDS. 
Unless otherwise specified, SoftDS is the default choice. 
Additionally, both training and inference can also be conducted using a simple AVG strategy.}

\begin{algorithm}
\caption{Calculate $\mathbf{\hat{y}}$}
\label{alg-main-ds}
\begin{algorithmic}[1]
\Require $\mathbf{\tilde{y}}$
\State $\mathbf{\hat{y}}_i \Leftarrow \frac{1}{K}\sum_{j=1}^K \mathbf{\tilde{y}}_{i,j}   $
\State $\boldsymbol{\pi}_j \Leftarrow (\frac{1}{C})_{C\times C}$
\State $iter \Leftarrow 0$
\While{$iter \leq maxiter$} 
    \State Update $\boldsymbol{\pi}$ according to \equationautorefname~\ref{eqa-main-updpi}
    \State Update $\mathbf{\hat{y}}$ according to \equationautorefname~\ref{eqa-main-updy-norm}
    \If{the update of $\pi$ is small enough}
        \State Break 
    \EndIf
    \State $iter \Leftarrow iter +1$ 
\EndWhile
\end{algorithmic}
\end{algorithm}

\paragraph{Dataset splitting}

Once we obtain the estimated true labels, i.e. $\mathbf{\hat{y}}$, we can calculate the entropy corresponding to each sample's prediction
\begin{equation}
entrop(\mathbf{x}_i)=-\sum_{j=1}^C\mathbf{\hat{y}}_{i,j}\cdot \log \mathbf{\hat{y}}_{i,j}.
\end{equation}
By sorting the samples based on their entropy values, those with higher entropy are considered to have lower model confidence and are classified as unlabeled samples. 
Conversely, those with lower entropy are regarded as having higher model confidence and are classified as labeled samples.
The division ratio is denoted as $\gamma$.
Thus, the original dataset is divided into a labeled dataset and an unlabeled dataset, $\mathcal{D}=\{\mathcal{D}_{labeled}, \mathcal{D}_{unlabled}\}$.



\begin{table*}[!htbp]
\centering
\caption{Results on PACS.}
\label{table-main-pacs-res}
\resizebox{.65\textwidth}{!}{
\begin{tabular}{lccccc}
\toprule
Domains     & A                   & C                   & P                   & S                   & AVG           \\ \midrule
ERM         & 77.5+/-1.8          & 72.0+/-1.4          & 93.9+/-0.8          & 63.3+/-2.2          & 76.7          \\
Mixup       & 81.0+/-1.1          & 71.8+/-0.4          & 94.9+/-0.7          & 64.1+/-0.7          & 77.9          \\
GroupDRO    & 80.8+/-2.7          & 71.3+/-2.2          & 93.8+/-1.6          & 61.9+/-2.2          & 77            \\
IRM         & 78.1+/-2.1          & 73.3+/-2.3          & 94.4+/-0.8          & 62.5+/-2.7          & 77.1          \\
VREx        & 79.9+/-2.4          & 74.0+/-1.8          & 96.0+/-0.2          & 60.2+/-2.1          & 77.5          \\
CORAL       & 79.0+/-1.5          & 71.0+/-1.8          & 95.8+/-1.4          & 62.2+/-2.0          & 77            \\
ADRMX       & 83.4+/-0.9          & 75.7+/-0.9          & 96.3+/-0.3          & 65.1+/-1.8          & 80.1          \\
ERM++ & 84.9+/-1.0          & 76.5+/-0.9          & 96.0+/-0.7          & 72.3+/-0.9          & 82.4          \\ \midrule
Ours        & \textbf{86.2+/-0.9} & \textbf{78.3+/-0.1} & \textbf{96.7+/-0.5} & \textbf{73.5+/-1.9} & \textbf{83.7} \\ \botrule
\end{tabular}
}
\end{table*}

\paragraph{Discussion}

During inference, we can either directly average the predictions from each hidden feature detector or treat them as different annotators and perform DS inference. 
Note that this is merely one possible implementation. 
In fact, our method is highly flexible. 
MixMatch can be replaced with other semi-supervised algorithms, such as SoftMatch~\citep{chensoftmatch}, and the aggregation of both training and inference methods can be further improved, e.g. adding class prior.

\section{Experiments}
\label{sec-exp}

We comprehensively evaluate our method across three image classification benchmarks. 
We follow the setting in \citep{qiaounderstanding} based on Domainbed~\citep{gulrajanisearch}.
The training data is randomly divided into two subsets, with $80\%$ used for training and $20\%$ for validation. 
For each task, we perform 20 hyperparameter search configurations, with each configuration evaluated across 3 independent experiments.
To ensure fairness, we re-implement seven state-of-the-art comparison methods: ERM, CORAL~\citep{sun2016deep}, Mixup~\citep{zhang2018mixup}, GroupDRO~\citep{sagawadistributionally}, IRM~\citep{arjovsky2019invariant}, VREx~\citep{krueger2021out}, ADRMX~\citep{demirel2023adrmx}, ERM++~\citep{teterwak2024ermimprovedbaselinedomain}\footnote{For fairness, we only utilize parts characters of the original version in our implementation.}. 

\begin{table*}[!htbp]
\caption{Results on VLCS.}
\centering
\label{tab-main-results-vlcs}
\resizebox{.65\textwidth}{!}{
\begin{tabular}{lccccc}
\toprule
Domains     & V                   & L                   & C                   & S                   & AVG           \\ \midrule
ERM         & 98.2+/-0.3          & 62.0+/-1.6          & 72.3+/-0.9          & 77.3+/-0.8          & 77.4          \\
Mixup       & 97.3+/-0.9          & 62.3+/-1.1          & 71.9+/-0.4          & 77.4+/-0.5          & 77.2          \\
GroupDRO    & 97.9+/-0.3          & 62.5+/-0.9          & \textbf{74.0+/-1.2} & 74.8+/-0.9          & 77.3          \\
IRM         & 98.0+/-0.7          & \textbf{63.8+/-0.4} & 68.9+/-1.1          & 75.0+/-2.1          & 76.4          \\
VREx        & 98.3+/-0.4          & \textbf{63.8+/-0.9} & 73.6+/-0.9          & 74.8+/-0.8          & 77.6          \\
CORAL       & 97.4+/-1.3          & 63.7+/-1.0          & 71.5+/-1.3          & 73.5+/-0.9          & 76.5          \\
ADRMX       & 98.5+/-0.2          & \textbf{63.8+/-1.4} & 71.5+/-0.9          & 74.6+/-1.4          & 77.1          \\
ERM++ & \textbf{99.0+/-0.2} & 60.0+/-0.4          & 72.9+/-0.4          & 78.4+/-0.5          & 77.6          \\ \midrule
Ours        & 98.9+/-0.3          & 62.4+/-0.6          & 73.6+/-0.8          & \textbf{79.9+/-0.8} & \textbf{78.7}\\ \botrule
\end{tabular}}
\end{table*}

\begin{table*}[!htbp]
\caption{Results on OfficeHome.}
\centering
\label{tab-main-results-off}
\resizebox{.65\textwidth}{!}{
\begin{tabular}{lccccc}
\toprule
Domains     & A                   & C                   & P                   & R                   & AVG           \\ \midrule
ERM         & 55.0+/-0.2          & 46.8+/-1.0          & 70.0+/-0.4          & 70.9+/-0.3          & 60.7          \\
Mixup       & 55.5+/-1.4          & 48.1+/-0.3          & 69.6+/-0.3          & 71.3+/-0.3          & 61.1          \\
GroupDRO    & 56.1+/-0.8          & 47.7+/-0.7          & 69.8+/-0.5          & 72.1+/-0.1          & 61.4          \\
IRM         & 52.6+/-0.3          & 42.9+/-0.3          & 67.2+/-0.7          & 67.5+/-1.1          & 57.6          \\
VREx        & 55.2+/-0.3          & 45.0+/-0.5          & 68.6+/-0.1          & 70.3+/-0.7          & 59.8          \\
CORAL       & 58.9+/-1.0          & 48.7+/-1.1          & 71.1+/-0.2          & 73.2+/-0.6          & 63            \\
ADRMX       & 57.7+/-0.4          & 46.8+/-0.8          & 70.0+/-0.1          & 72.6+/-0.4          & 61.8          \\
ERM++ & 61.3+/-0.6          & 53.1+/-0.1          & 74.7+/-0.2          & 76.0+/-0.3          & 66.3          \\ \midrule
Ours        & \textbf{62.0+/-0.2} & \textbf{55.2+/-0.4} & \textbf{75.7+/-0.3} & \textbf{77.0+/-0.3} & \textbf{67.5} \\ \botrule
\end{tabular}}
\end{table*}

\subsection{PACS}

\paragraph{Datasets}

PACS~\citep{li2017deeper} is a benchmark dataset for classification, consisting of four domains: photo, art painting, cartoon, and sketch.
There are significant differences in image styles across these domains.
The dataset includes seven classes with a total of 9,991 images.
Following~\citep{qiaounderstanding}, we randomly flip $25\%$ of the labels.

\paragraph{Results}

The results are shown in \tableautorefname~\ref{table-main-pacs-res}, and we have the following observations:
1) Our method consistently improves performance over all baseline methods and achieves the best results in almost every domain, with the improvement about 7 percentage points compared to ERM.
2) On the PACS dataset, which exhibits significant style variations, most DG methods outperform ERM. However, in certain challenging domains like C and S, ERM performs better than methods such as CORAL. ERM++, based on different training strategies, shows consistent improvements over ERM, and our method further enhances ERM++. 
3) Considering domain-specific features has a notable impact on both DG and noise-robust methods, with ADRMX outperforming methods like CORAL.

\begin{table*}[h]
\caption{Results without using semi-supervision and iteration or without using Dawid-Skene training.}
\centering
\label{tab-main-ablat}
\resizebox{.65\textwidth}{!}{
\begin{tabular}{lccccc}
\toprule
Domains &A&C&P&S&AVG\\ \midrule
Base       & 77.5+/-1.8 & 72.0+/-1.4 & 93.9+/-0.8 & 63.3+/-2.2 & 76.7 \\ \midrule
Linear+AVG & 78.2+/-0.4 & 69.9+/-1.1 & 91.9+/-0.5 & 63.1+/-1.5 & 75.8 \\
Linear+DS  & 79.7+/-1.5 & 76.0+/-1.0 & 95.0+/-0.6 & 69.8+/-1.1 & 80.1 \\
LSTM+AVG   & 79.6+/-1.4 & 69.1+/-1.5 & 93.8+/-0.9 & 65.4+/-2.4 & 77   \\
LSTM+DS    & 79.5+/-1.8 & 72.8+/-1.1 & 94.9+/-0.7 & 68.1+/-1.3 & 78.8 \\ \midrule
AVG+AVG & 78.2+/-1.3 & 69.9+/-0.7 & 90.5+/-0.6 & 64.0+/-3.3 & 75.7 \\
AVG+DS  & 82.0+/-1.5 & 77.0+/-0.2 & 95.1+/-0.9 & 70.7+/-1.1 & 81.2 \\ \midrule
Ours    & 81.8+/-1.5 & 76.7+/-0.1 & 95.4+/-0.6 & 70.4+/-1.0 & 81  \\ \botrule
\end{tabular}
}
\end{table*}

\begin{figure*}[!htbp]
    \centering
    \subfigure[$K$]{
        \includegraphics[height=0.22\textwidth]{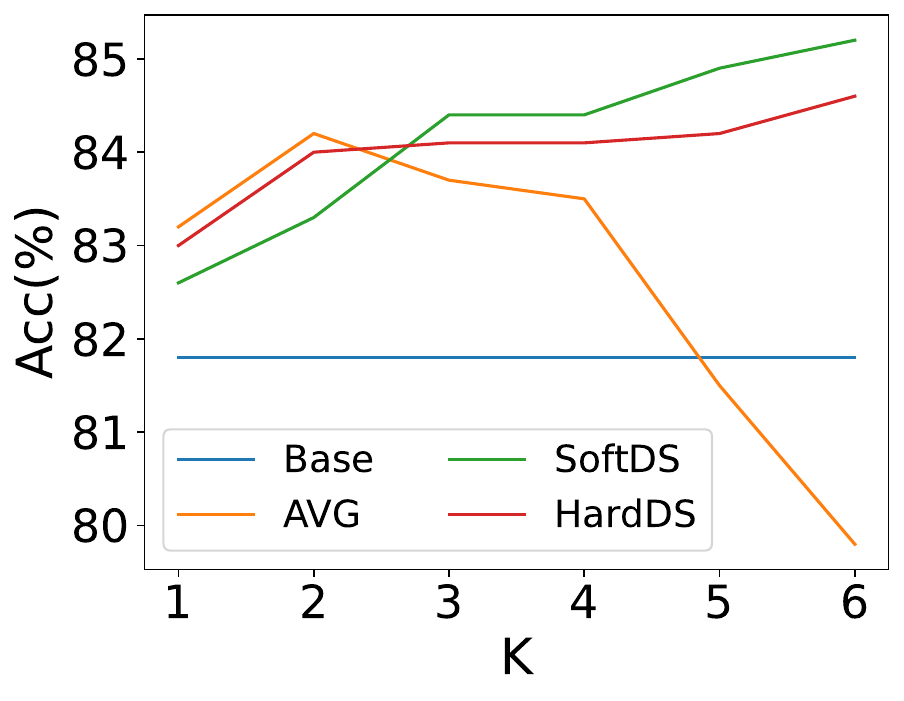}
        \label{fig-sens-k}
    }
    \subfigure[Round]{
        \includegraphics[height=0.22\textwidth]{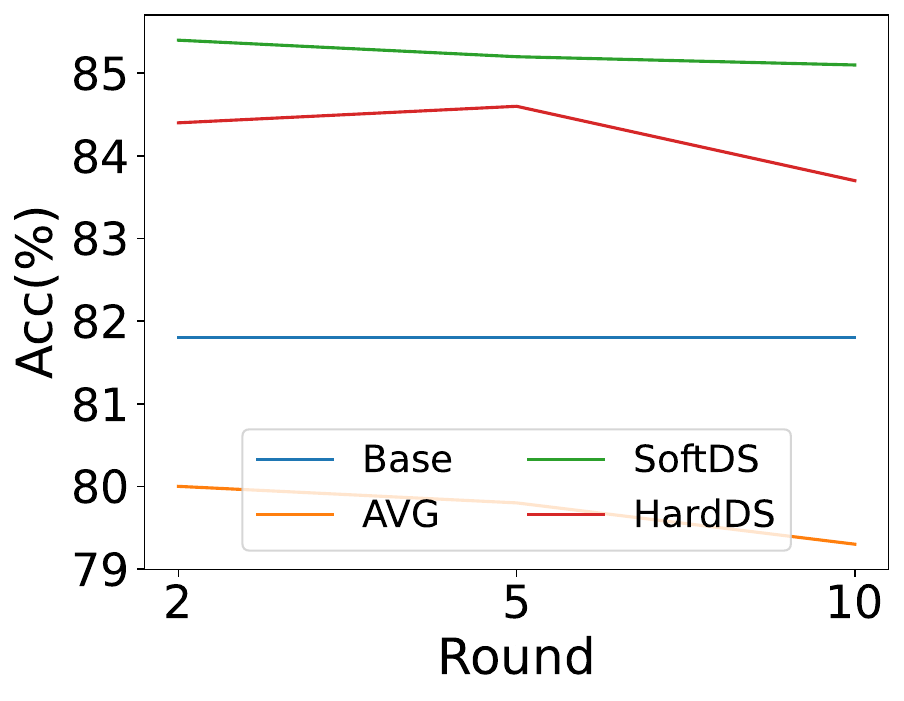}
        \label{fig-sens-round}
    }
    \subfigure[Noise Rate ($\gamma$)]{
        \includegraphics[height=0.22\textwidth]{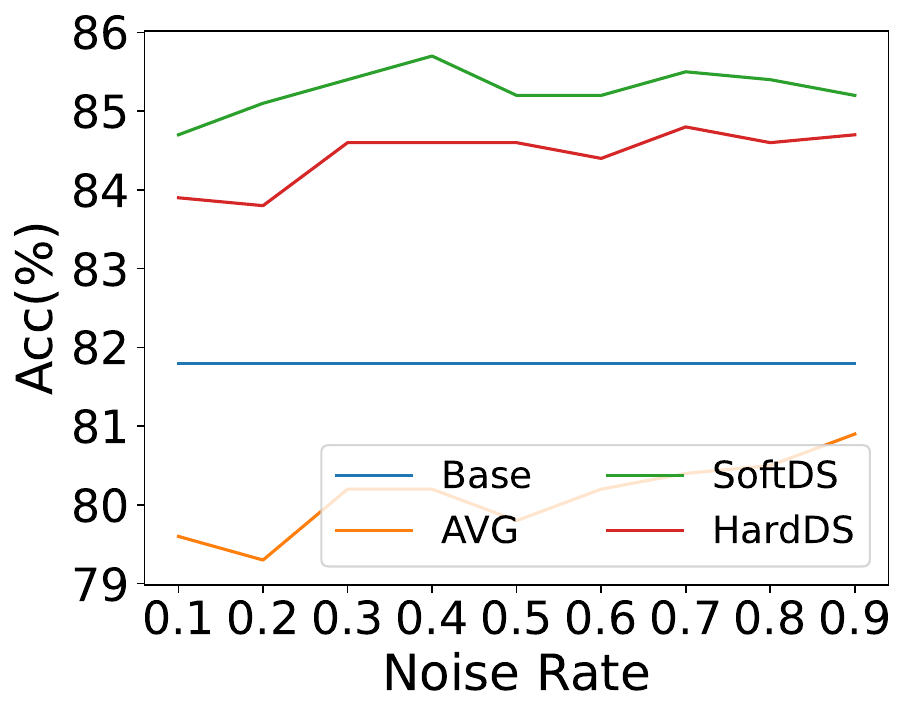}
        \label{fig-sensrate}
    }
    \caption{HyperParameter Sensitivity Analysis.}
    \label{fig-main-sens}
\end{figure*}

\subsection{VLCS}
\paragraph{Datasets}
\textbf{VLCS}~\citep{fang2013unbiased} consists of four photographic domains: Caltech101, LabelMe, SUN09, and VOC2007.
The dataset includes 10,729 samples spanning five classes.
One domain is designated as the test domain, which remains unseen during training, while the remaining domains are used for training.
Here, we use hard labels for the final integration, e.g. HardDS for inference.

\paragraph{Results}

The results for VLCS are shown in \tableautorefname~\ref{tab-main-results-vlcs}, and we summarize our key observations as follows:
1) Our method, built upon the well-performing ERM++, achieves further improvements, delivering the best overall performance. 
2) Noise does not always have a negative impact. For example, on VLCS, adding noise yields results comparable to the clean setting in DomainBed. However, our method consistently enhances performance.
3) No single method is always the best. For instance, in the V domain of VLCS, methods like IRM perform better, while in the C domain, GroupDRO achieves superior results. This highlights the need to choose appropriate methods based on the dataset characteristics.

\subsection{OfficeHome}
\paragraph{Datasets}
Similar to PACS, OfficeHome~\citep{venkateswara2017deep} also consists of four distinct domains: Art, Clipart, Product, and Real\_World.
The dataset contains a total of 15,588 images but includes a larger number of categories, with 65 classes.

\paragraph{Results}

The results for OfficeHome are shown in \tableautorefname~\ref{tab-main-results-off}, and we summarize our key observations as follows:
1) Our method, built upon the well-performing ERM++, achieves further improvements on both datasets, delivering the best overall performance. It even achieves the highest accuracy in every domain.
2) The degree of improvement varies across different datasets due to differences in internal hidden features. Therefore, both feature probing strategies and inference ensemble strategies must be carefully selected.

\begin{figure}[!t]
\centering
\includegraphics[width=0.48\textwidth]{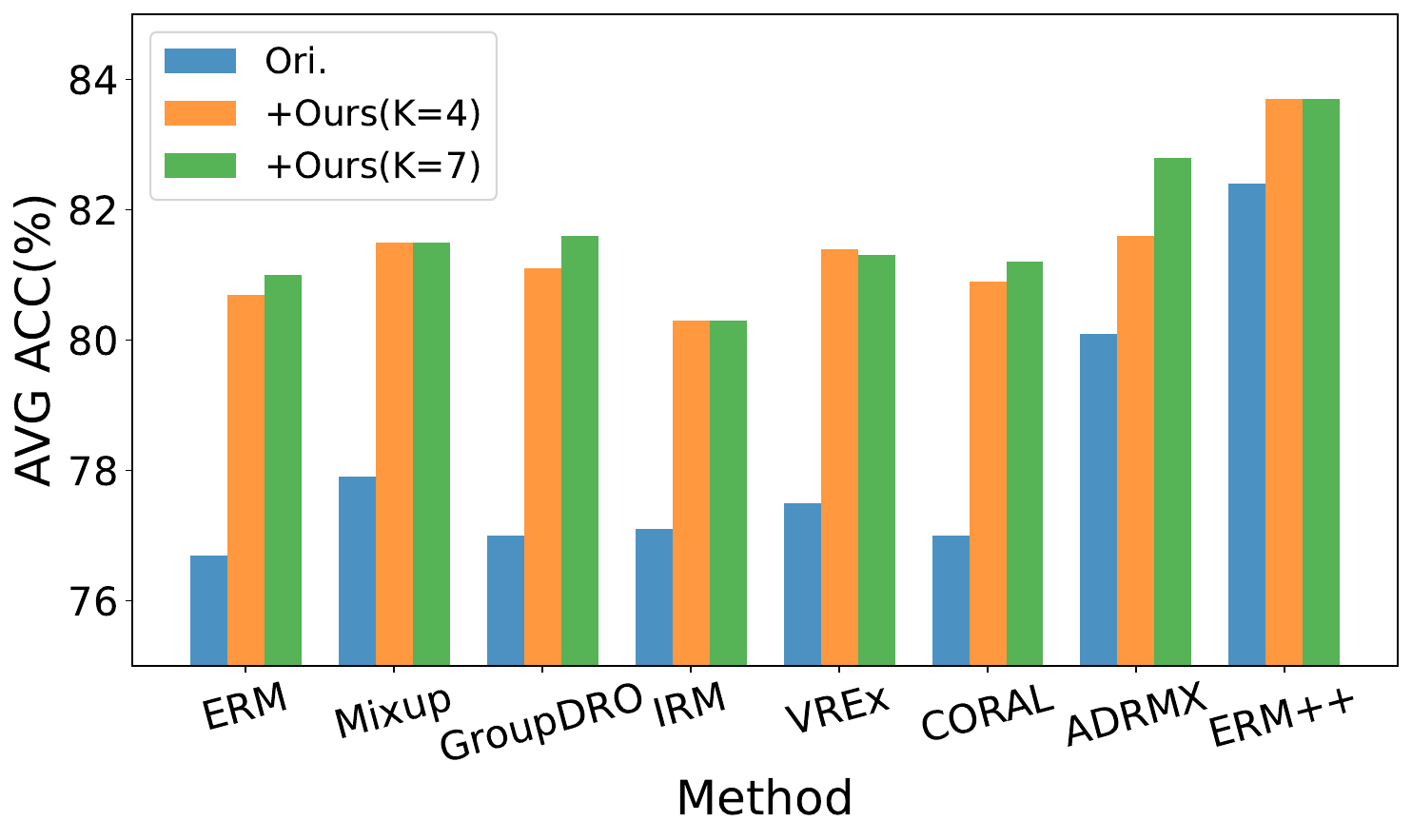}
\caption{Universal results on PACS.}\label{fig-main-discuss-univ}
\end{figure}


\begin{figure*}[!htbp]
    \centering
    \hspace{-.2in}
    \subfigure[Extensibility on VLCS.]{
        \includegraphics[height=0.18\textwidth]{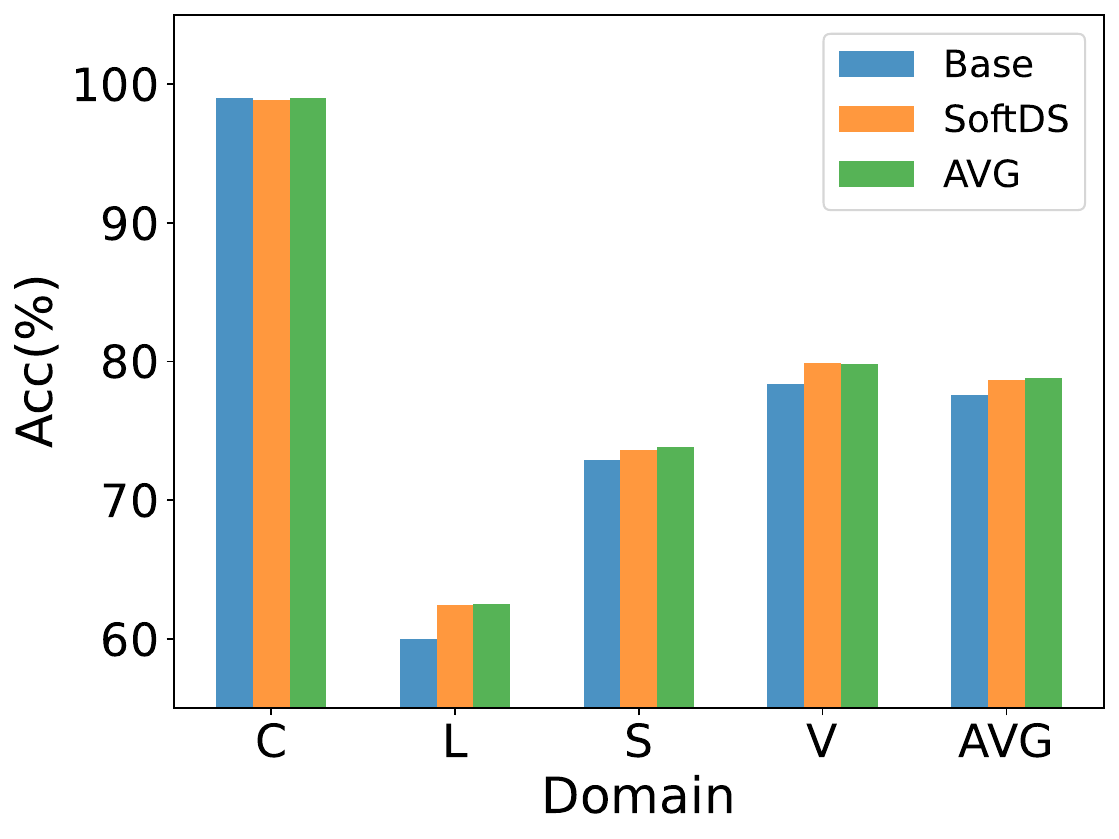}
        \label{fig-diss1-vlcs}
    }
    \subfigure[Exten. on OfficeHome.]{
        \includegraphics[height=0.18\textwidth]{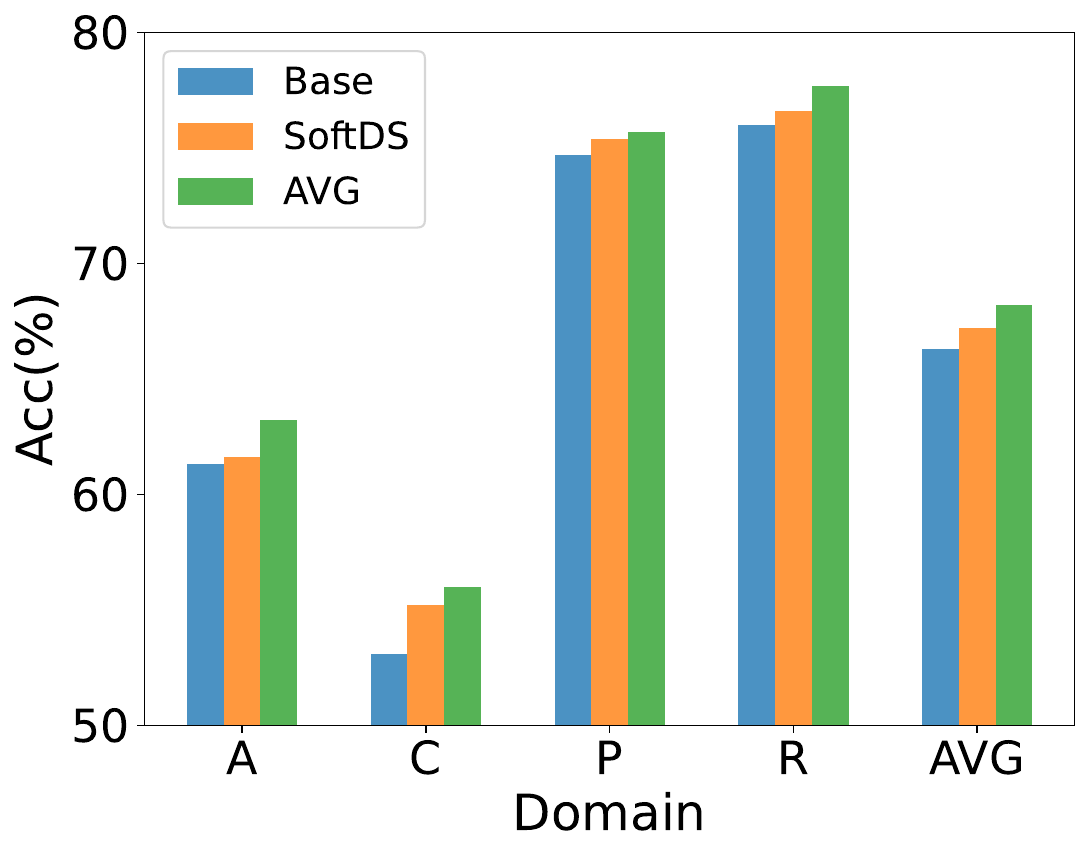}
        \label{fig-diss1-off}
    }
    \subfigure[Results with no fea. training.]{
        \includegraphics[height=0.18\textwidth]{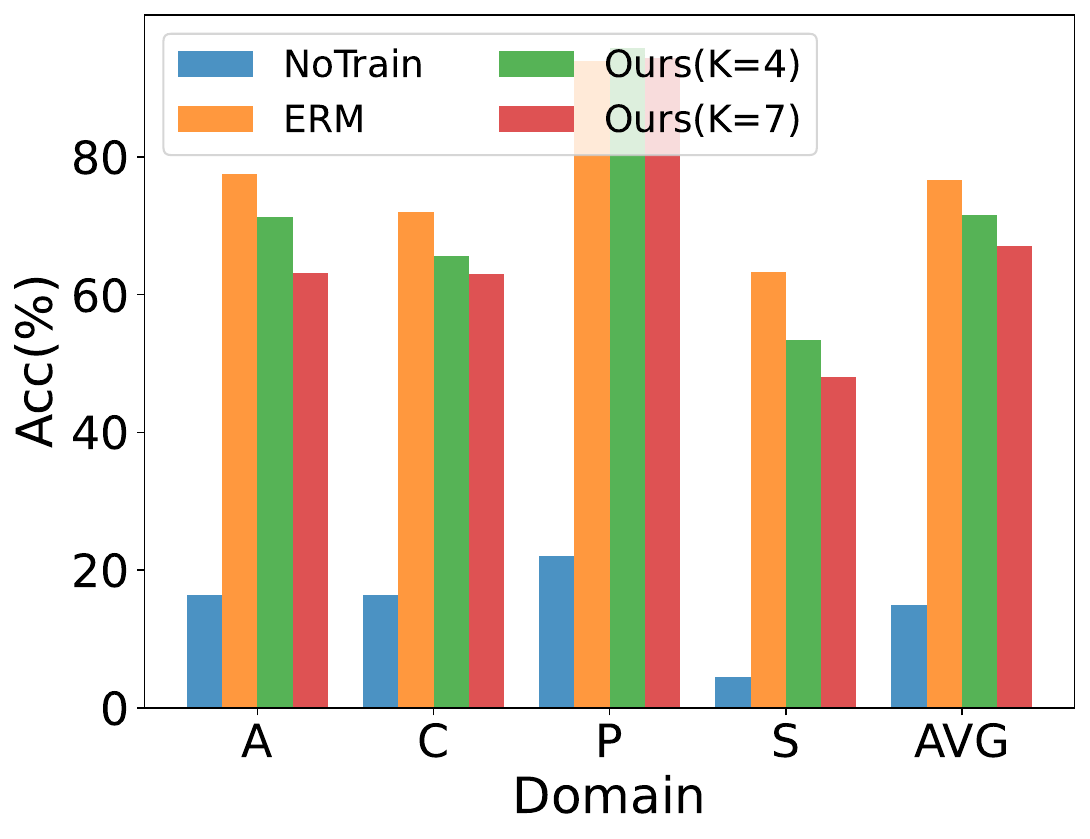}
        \label{fig-diss-notrain}
    }
    \subfigure[Noise Rate. (Setting)]{
        \includegraphics[height=0.18\textwidth]{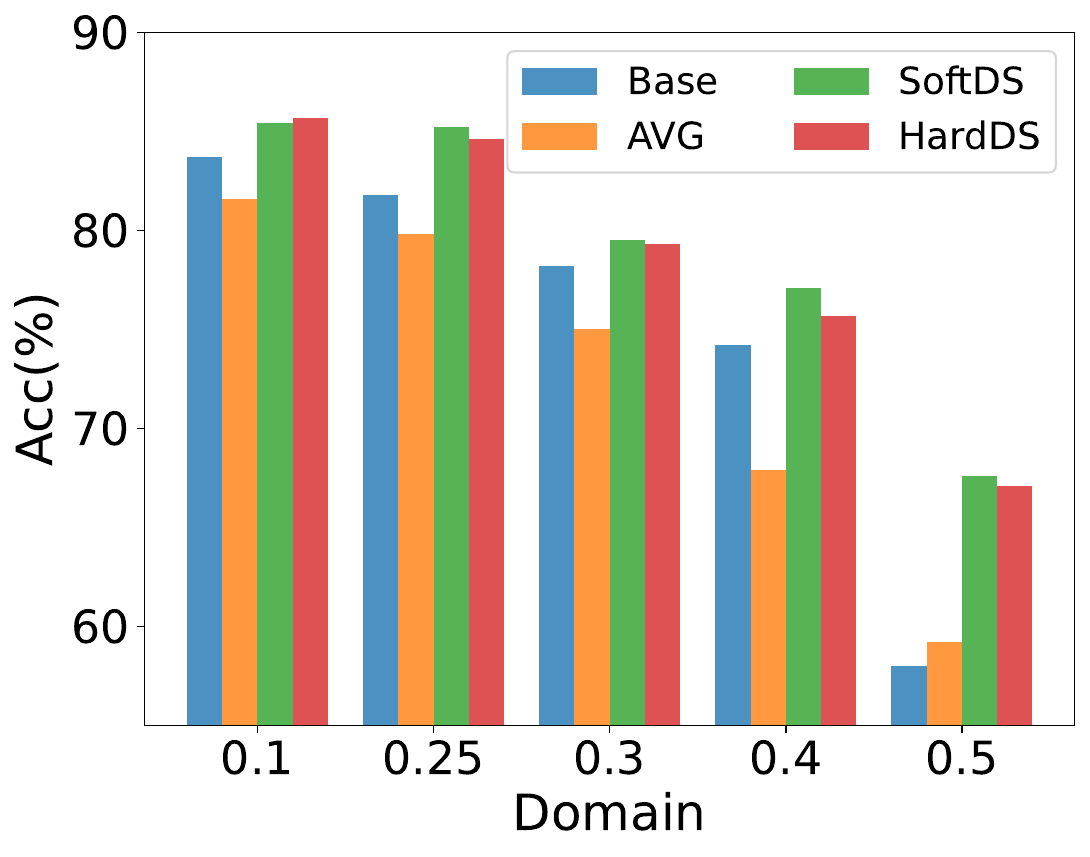}
        \label{fig-diss-rate}
    }
    \caption{More discussion. }
    \label{fig-main-dissc}
\end{figure*}

\subsection{Analysis}


\subsubsection{Ablation Study}



\tableautorefname~\ref{tab-main-ablat} presents the results of the ablation study, demonstrating that each module in our method contributes to performance improvement.
\footnote{In \tableautorefname~\ref{tab-main-ablat}, if there is only one AVG/DS, it means that noise splitting is not applied (in this situation, we directly train probing classifiers instead of applying semi-supervised learning), and different strategies are used during inference.
If there are two AVG/DS, the first one represents the strategy used during training, while the second one indicates the strategy used during inference.}
Additionally, we observe that replacing certain module implementations within the method can sometimes lead to better results, highlighting the need to select appropriate implementations based on the characteristics of the data and model.
Furthermore, we have the following observations:
1) Directly averaging the results from six hidden feature probing layers may degrade performance. This aligns with \figureautorefname~\ref{fig-mot2}, as earlier features have lower task relevance.
2) Replacing the linear probing layer with an LSTM does not yield significant improvements, suggesting that feature quality has a greater impact.
3) Both the semi-supervised approach and the DS algorithm lead to noticeable performance gains.

\subsubsection{Hyperparameter Sensitivity}

We also conducted a hyperparameter sensitivity analysis on the following three hyperparameters, with results shown in \figureautorefname~\ref{fig-main-sens}. Our key observations are as follows:
1) Our method demonstrates relative stability across different hyperparameters.
2) Both SoftDS and HardDS inference outperform ERM, while the AVG inference approach performs worse due to the influence of the early-layer hidden feature probing classifiers.
3) The number of iterations impacts performance, but more iterations do not necessarily lead to better results. This suggests that selecting an appropriate stopping point is crucial.
4) The choice of noise ratio affects performance. In our experiments, we set the noise ratio to 0.25, and our sensitivity analysis indicates that values close to 0.25 tend to yield better results. Interestingly, for SoftDS, a noise ratio of around 0.4 also achieves good performance.


\begin{table*}[!htbp]
\caption{Results on Skin Cancer.}
\centering
\label{tab-main-app1}
\resizebox{.65\textwidth}{!}{
\begin{tabular}{lccccc}
\toprule
Domains     & Group0              & Group1              & Group2              & Group3              & AVG           \\ \midrule
ERM         & 67.8+/-0.7          & 73.7+/-1.4          & 61.5+/-1.9          & 60.5+/-1.1          & 65.9          \\
Mixup       & 67.3+/-0.8          & 71.9+/-0.8          & 59.7+/-2.5          & 60.1+/-0.2          & 64.7          \\
GroupDRO    & 66.8+/-0.5          & 73.7+/-0.1          & 59.5+/-0.5          & 57.0+/-0.6          & 64.3          \\
IRM         & 66.8+/-1.4          & 73.8+/-0.3          & 63.5+/-1.9          & 54.6+/-2.0          & 64.6          \\
VREx        & 68.5+/-1.4          & 73.3+/-0.7          & 62.4+/-0.5          & 57.0+/-2.0          & 65.3          \\
CORAL       & 67.0+/-0.8          & 71.6+/-1.1          & 61.5+/-0.7          & 56.8+/-2.7          & 64.2          \\
ADRMX       & 70.0+/-0.4          & 71.0+/-0.3          & 60.1+/-0.1          & 57.5+/-1.6          & 64.6          \\
ERM++ & 69.1+/-0.6          & 75.0+/-0.7          & 65.0+/-0.8          & 59.8+/-1.4          & 67.2          \\ \midrule
Ours        & \textbf{71.2+/-0.5} & \textbf{75.8+/-0.8} & \textbf{69.1+/-0.7} & \textbf{61.7+/-1.1} & \textbf{69.5} \\ \botrule
\end{tabular}
}
\end{table*}

\subsubsection{More Discussion}

In this section, we discuss the extensibility and robustness of our method.

\paragraph{Is our method universally effective?} 
Yes, as a post-processing approach, our method is generally effective across various DG methods, enhancing their performance in noisy DG scenarios. 
From \figureautorefname~\ref{fig-main-discuss-univ}, we observe the following:
1) On the PACS dataset, our method improves the performance of all baseline methods, with the highest improvement exceeding 4 percentage points.
2) The performance gain of our method is somewhat influenced by the underlying baseline. Generally, the better the baseline performs, the better our method performs, highlighting the importance of strong underlying feature representations.
3) On PACS, using six layers of hidden features for probing typically yields better results. However, this is not always the case, as seen with VREx. This suggests that incorporating model understanding into parameter selection could further enhance the performance of our approach.

\paragraph{What happens if we modify certain components of \framework?} 
To explore this, we replaced SoftDS with AVG during the training process while keeping the aggregation method in the inference stage unchanged. 
As shown in \figureautorefname~\ref{fig-diss1-vlcs} and \figureautorefname~\ref{fig-diss1-off}, the results on VLCS and OfficeHome indicate that using AVG can even lead to better performance (PACS results are shown in \tableautorefname~\ref{tab-main-ablat}.). 
On OfficeHome, the average performance improved by approximately one percentage point (with $1.9\%$ compared to ERM++). 
This demonstrates the strong adaptability of our method selecting an appropriate implementation can yield even better results.

\paragraph{Can our method improve an untrained model?} 
As shown in \figureautorefname~\ref{fig-diss-notrain}, our approach also enhances the performance of models that have not undergone task-related training. 
Without any training, the model produces nearly random outputs, with an average accuracy of $14.9\%$. 
However, using our post-processing method without modifying the model's backbone, the accuracy improves to $71.5\%$, approaching the performance of ERM ($76.7\%$). 
This suggests that even an untrained model can extract some task-relevant features. 
Additionally, we observe that using too many hidden feature detectors in this scenario is not beneficial, indicating that shallow hidden features are relatively harder to leverage effectively.

\begin{table*}[!htbp]
\caption{Results on Organ classification.}
\centering
\label{tab-main-app2}
\resizebox{.6\textwidth}{!}{
\begin{tabular}{lcccc}
\toprule
Dataset     & A                   & C                   & S                   & AVG           \\ \midrule
ERM         & 75.3+/-1.5          & 79.8+/-0.5          & 52.8+/-1.1          & 69.3          \\
Mixup       & 74.3+/-0.4          & 75.5+/-0.8          & 54.9+/-1.1          & 68.3          \\
GroupDRO    & 75.9+/-1.3          & 76.6+/-0.5          & 53.7+/-1.1          & 68.7          \\
IRM         & 74.8+/-1.0          & 77.5+/-1.7          & 54.1+/-1.2          & 68.8          \\
VREx        & 72.3+/-3.4          & 75.7+/-1.0          & 53.9+/-1.0          & 67.3          \\
CORAL       & 74.2+/-1.2          & 77.0+/-0.2          & 52.1+/-1.9          & 67.8          \\
ADRMX       & 72.3+/-1.0          & 74.6+/-1.0          & 53.0+/-0.5          & 66.6          \\
ERM++ & 80.5+/-0.5          & 79.9+/-0.4          & 57.9+/-0.4          & 72.8          \\ \midrule
Ours        & \textbf{83.8+/-0.6} & \textbf{84.0+/-0.7} & \textbf{58.8+/-0.9} & \textbf{75.5} \\ 
\botrule
\end{tabular}
}
\end{table*}

\paragraph{Is our method still effective under different noise ratios?}
In previous experiments, we primarily set the noise ratio to $0.25$. 
Here, we evaluate our method under different noise levels. 
As shown in \figureautorefname~\ref{fig-diss-rate}, as the noise ratio increases, the performance of all methods gradually declines. 
However, our approach continues to provide improvements. 
Notably, even when the noise ratio reaches $0.5$, our method still achieves an improvement of nearly $10\%$ over the baseline. 
This demonstrates the robustness of our approach.

\section{Application on Real Datasets}\label{sec-app}

\subsection{Skin Cancer}

Skin cancer is one of the most common types of cancer worldwide, with its incidence rising due to increased exposure to ultraviolet radiation. 
Early detection and accurate diagnosis are crucial, as timely treatment can significantly improve survival rates. 
Advancing research in skin cancer can lead to better diagnostic tools, more effective treatments, and a deeper understanding of its underlying causes~\citep{jerant2000early}.

\paragraph{Dataset and settings}

We use the HAM10000 dataset~\citep{tschandl2018ham10000}, which contains $10,015$ dermatoscopic images across seven categories. 
After filtering out samples without age records, we divide the dataset into four domains based on age. 
Since some categories have few samples, we merge these into a single "other" category, resulting in four final categories: bkl, mel, nv, and others. 
Additionally, since the nv category has an excessive number of samples, we randomly sample 600 images per domain from this category.
This results in a dataset containing $5,698$ images across four domains and four categories.
In the training data, we randomly flip $25\%$ of the labels.
\footnote{Our current setting does not account for the impact of class imbalance (as our DS inference assumes a uniform class distribution). Notably, Group0 exhibits severe class imbalance, so we use AVG inference for this group. In future extensions, we can incorporate class priors into DS inference to better handle such imbalances.}

\paragraph{Results}

From \tableautorefname~\ref{tab-main-app1}, we can see that \framework also performs well in real-world applications, improving the average accuracy of the second-best method, ERM++, by $2.3\%$ (noting that ERM++ only improves $1.3\%$ over ERM), and achieving an overall $3.6\%$ improvement compared to ERM. 
This demonstrates the potential of \framework for practical applications. 
Moreover, selecting an appropriate implementation can further enhance our approach, suggesting that automated selection could be an interesting direction for future research.

\subsection{Organ Classification}

\paragraph{Dataset and settings}
In this section, we focus on organ recognition and classification, which plays a crucial role in medical imaging. 
We demonstrate the performance of our method using the Organ\{A,C,S\} datasets from the MedMnist benchmark~\citep{medmnistv1,medmnistv2}. 
The new version of the MedMnist dataset, MedMNIST+, contains $224\times224$ 2D images. 
The Organ\{A,C,S\}MNIST is based on 3D computed tomography (CT) images from the Liver Tumor Segmentation Benchmark (LiTS)~\citep{bilic2023liver}, representing the Axial, Coronal, and Sagittal views. 
We treat each subset as a domain. 
The dataset contains a total of 11 categories and nearly 100,000 images. 
To add complexity while reducing training cost, we randomly select 200 images for each category in every domain, totaling 3 domains, 11 categories, and 6600 images. Additionally, we randomly flip $25\%$ of the class labels.
Since the basic classification accuracy for the last domain is below $60\%$, which leads to weaker discrimination ability in the shallow features, we set $K=4$ for the last domain, while setting $K=7$ for the other domains.

\paragraph{Results}

As shown in \tableautorefname~\ref{tab-main-app2}, our method outperforms the second-best method, ERM++, by $2.7\%$ and the baseline method by $6.2\%$, further demonstrating the potential of our approach in real-world applications. 
Additionally, it is surprising that some methods designed for out-of-distribution (OOD) scenarios perform even worse than the baseline algorithm in this real-world application. 
This highlights that different methods have their own applicability depending on the specific application or data. 
In most cases, selecting the appropriate implementation and parameters can lead to stable improvements with our method.

\section{Limitation}\label{sec-limit}

Since this work represents the first exploration of post-processing through model self-ensemble in the field of noisy domain generalization, there are still some limitations that warrant improvement:
\begin{enumerate}
    \item \textbf{Larger Models, More Implementation Methods, and Broader Applications}: The current approach is primarily based on the Domainbed library~\citep{gulrajanisearch} and focuses on ResNet50. Future work could explore Transformer-based models~\citep{DosovitskiyB0WZ21} and alternative semi-supervised implementations, such as SoftMatch~\citep{chensoftmatch}.
    \item \textbf{Towards an End-to-End Method}: The current method requires aggregating multiple predictions for ensemble learning, which has not yet formed a unified end-to-end method. Future improvements could consider adopting approaches similar to DL-CL~\citep{rodrigues2018deep} for a more integrated solution.
    \item \textbf{Further Refinement and Optimization}: While the current approach demonstrates promising results, more meticulous hyperparameter tuning could yield even better performance. Additionally, some implementation details and methods could be further improved and refined. Automated selection could be an interesting research topic.
\end{enumerate}

\section{Conclusion}\label{sec-conclu}

In this paper, we propose a general and extensible post-processing method, i.e. \framework, based on model self-ensemble to address the problem of noisy domain generalization. 
\framework consists of two parts, i.e. probing classifier training and prediction ensemble inference.
It iteratively identifies noisy data and achieves robust ensemble results. 
Through extensive experiments and two real-world applications, we demonstrate that our method can make existing OOD methods rework. 
Additionally, we find that \framework is highly flexible and it can still provide certain benefits even for untrained models, offering insights for future model utilization and exploration.

\section*{Declarations}

\paragraph{Data availability}
The datasets analyzed during the current study are available in Domainbed~\citep{gulrajanisearch}(\url{https://github.com/facebookresearch/DomainBed}), Skin Cancer~\citep{tschandl2018ham10000}
(\url{https://www.kaggle.com/datasets/farjanakabirsamanta/skin-cancer-dataset}), and MedMnist~\citep{medmnistv1,medmnistv2} (\url{https://zenodo.org/records/10519652}). 
\paragraph{Code availability}
We plan to open source code for the community in the near future.

\bibliography{sn-bibliography}

@article{delussu2024synthetic,
  title={Synthetic Data for Video Surveillance Applications of Computer Vision: A Review},
  author={Delussu, Rita and Putzu, Lorenzo and Fumera, Giorgio},
  journal={International Journal of Computer Vision},
  pages={1--37},
  year={2024},
  publisher={Springer}
}

@inproceedings{jianggraphcare,
  title={GraphCare: Enhancing Healthcare Predictions with Personalized Knowledge Graphs},
  author={Jiang, Pengcheng and Xiao, Cao and Cross, Adam Richard and Sun, Jimeng},
  booktitle={The Twelfth International Conference on Learning Representations},
  year={2024}
}

@inproceedings{baek2024unexplored,
  title={Unexplored Faces of Robustness and Out-of-Distribution: Covariate Shifts in Environment and Sensor Domains},
  author={Baek, Eunsu and Park, Keondo and Kim, Jiyoon and Kim, Hyung-Sin},
  booktitle={Proceedings of the IEEE/CVF Conference on Computer Vision and Pattern Recognition},
  pages={22294--22303},
  year={2024}
}

@article{zhou2022domain,
  title={Domain generalization: A survey},
  author={Zhou, Kaiyang and Liu, Ziwei and Qiao, Yu and Xiang, Tao and Loy, Chen Change},
  journal={IEEE Transactions on Pattern Analysis and Machine Intelligence},
  volume={45},
  number={4},
  pages={4396--4415},
  year={2022},
  publisher={IEEE}
}

@article{song2022learning,
  title={Learning from noisy labels with deep neural networks: A survey},
  author={Song, Hwanjun and Kim, Minseok and Park, Dongmin and Shin, Yooju and Lee, Jae-Gil},
  journal={IEEE transactions on neural networks and learning systems},
  volume={34},
  number={11},
  pages={8135--8153},
  year={2022},
  publisher={IEEE}
}

@article{tschandl2018ham10000,
  title={The HAM10000 dataset, a large collection of multi-source dermatoscopic images of common pigmented skin lesions},
  author={Tschandl, Philipp and Rosendahl, Cliff and Kittler, Harald},
  journal={Scientific data},
  volume={5},
  number={1},
  pages={1--9},
  year={2018},
  publisher={Nature Publishing Group}
}

@inproceedings{li2017deeper,
  title={Deeper, broader and artier domain generalization},
  author={Li, Da and Yang, Yongxin and Song, Yi-Zhe and Hospedales, Timothy M},
  booktitle={Proceedings of the IEEE international conference on computer vision},
  pages={5542--5550},
  year={2017}
}

@article{wang2022generalizing,
  title={Generalizing to unseen domains: A survey on domain generalization},
  author={Wang, Jindong and Lan, Cuiling and Liu, Chang and Ouyang, Yidong and Qin, Tao and Lu, Wang and Chen, Yiqiang and Zeng, Wenjun and Philip, S Yu},
  journal={IEEE transactions on knowledge and data engineering},
  volume={35},
  number={8},
  pages={8052--8072},
  year={2022},
  publisher={IEEE}
}

@inproceedings{sun2016deep,
  title={Deep coral: Correlation alignment for deep domain adaptation},
  author={Sun, Baochen and Saenko, Kate},
  booktitle={Computer Vision--ECCV 2016 Workshops: Amsterdam, The Netherlands, October 8-10 and 15-16, 2016, Proceedings, Part III 14},
  pages={443--450},
  year={2016},
  organization={Springer}
}

@inproceedings{rame2023model,
  title={Model ratatouille: Recycling diverse models for out-of-distribution generalization},
  author={Ram{\'e}, Alexandre and Ahuja, Kartik and Zhang, Jianyu and Cord, Matthieu and Bottou, L{\'e}on and Lopez-Paz, David},
  booktitle={International Conference on Machine Learning},
  pages={28656--28679},
  year={2023},
  organization={PMLR}
}

@inproceedings{feng2023ot,
  title={Ot-filter: An optimal transport filter for learning with noisy labels},
  author={Feng, Chuanwen and Ren, Yilong and Xie, Xike},
  booktitle={Proceedings of the IEEE/CVF Conference on Computer Vision and Pattern Recognition},
  pages={16164--16174},
  year={2023}
}

@article{berthelot2019mixmatch,
  title={Mixmatch: A holistic approach to semi-supervised learning},
  author={Berthelot, David and Carlini, Nicholas and Goodfellow, Ian and Papernot, Nicolas and Oliver, Avital and Raffel, Colin A},
  journal={Advances in neural information processing systems},
  volume={32},
  year={2019}
}

@article{dawid1979maximum,
  title={Maximum likelihood estimation of observer error-rates using the EM algorithm},
  author={Dawid, Alexander Philip and Skene, Allan M},
  journal={Journal of the Royal Statistical Society: Series C (Applied Statistics)},
  volume={28},
  number={1},
  pages={20--28},
  year={1979},
  publisher={Wiley Online Library}
}

@article{pan2009survey,
  title={A survey on transfer learning},
  author={Pan, Sinno Jialin and Yang, Qiang},
  journal={IEEE Transactions on knowledge and data engineering},
  volume={22},
  number={10},
  pages={1345--1359},
  year={2009},
  publisher={IEEE}
}

@article{wang2018deep,
  title={Deep visual domain adaptation: A survey},
  author={Wang, Mei and Deng, Weihong},
  journal={Neurocomputing},
  volume={312},
  pages={135--153},
  year={2018},
  publisher={Elsevier}
}

@inproceedings{zhang2018mixup,
	title={mixup: Beyond Empirical Risk Minimization},
	author={Zhang, Hongyi and Cisse, Moustapha and Dauphin, Yann N and Lopez-Paz, David},
	booktitle={International Conference on Learning Representations},
	year={2018}
}

@inproceedings{yao2022improving,
  title={Improving out-of-distribution robustness via selective augmentation},
  author={Yao, Huaxiu and Wang, Yu and Li, Sai and Zhang, Linjun and Liang, Weixin and Zou, James and Finn, Chelsea},
  booktitle={International Conference on Machine Learning},
  pages={25407--25437},
  year={2022},
  organization={PMLR}
}

@article{li2024beyond,
  title={Beyond Finite Data: Towards Data-free Out-of-distribution Generalization via Extrapola},
  author={Li, Yijiang and Ren, Sucheng and Deng, Weipeng and Xu, Yuzhi and Gao, Ying and Ngai, Edith and Wang, Haohan},
  journal={arXiv preprint arXiv:2403.05523},
  year={2024}
}

@article{demirel2023adrmx,
  title={ADRMX: Additive Disentanglement of Domain Features with Remix Loss},
  author={Demirel, Berker and Aptoula, Erchan and Ozkan, Huseyin},
  journal={arXiv preprint arXiv:2308.06624},
  year={2023}
}

@inproceedings{zhang2021deep,
  title={Deep stable learning for out-of-distribution generalization},
  author={Zhang, Xingxuan and Cui, Peng and Xu, Renzhe and Zhou, Linjun and He, Yue and Shen, Zheyan},
  booktitle={Proceedings of the IEEE/CVF Conference on Computer Vision and Pattern Recognition},
  pages={5372--5382},
  year={2021}
}

@inproceedings{yangmanydg,
  title={ManyDG: Many-domain Generalization for Healthcare Applications},
  author={Yang, Chaoqi and Westover, M Brandon and Sun, Jimeng},
  year={2023},
  booktitle={The Eleventh International Conference on Learning Representations}
}

@inproceedings{yuclipceil,
  title={CLIPCEIL: Domain Generalization through CLIP via Channel rEfinement and Image-text aLignment},
  author={Yu, Xi and Yoo, Shinjae and Lin, Yuewei},
  year={2024},
  booktitle={The Thirty-eighth Annual Conference on Neural Information Processing Systems}
}

@inproceedings{rame2022fishr,
  title={Fishr: Invariant gradient variances for out-of-distribution generalization},
  author={Rame, Alexandre and Dancette, Corentin and Cord, Matthieu},
  booktitle={International Conference on Machine Learning},
  pages={18347--18377},
  year={2022},
  organization={PMLR}
}

@article{rame2022diverse,
  title={Diverse weight averaging for out-of-distribution generalization},
  author={Rame, Alexandre and Kirchmeyer, Matthieu and Rahier, Thibaud and Rakotomamonjy, Alain and Gallinari, Patrick and Cord, Matthieu},
  journal={Advances in Neural Information Processing Systems},
  volume={35},
  pages={10821--10836},
  year={2022}
}

@inproceedings{sunself,
  title={Self-cognitive Denoising in the Presence of Multiple Noisy Label Sources},
  author={Sun, Yi-Xuan and Zhang, Ya-Lin and Han, Bin and Li, Longfei and Zhou, Jun},
  year={2024},
  booktitle={Forty-first International Conference on Machine Learning}
}

@inproceedings{kim2024learning,
  title={Learning with Structural Labels for Learning with Noisy Labels},
  author={Kim, Noo-ri and Lee, Jin-Seop and Lee, Jee-Hyong},
  booktitle={Proceedings of the IEEE/CVF Conference on Computer Vision and Pattern Recognition},
  pages={27610--27620},
  year={2024}
}

@inproceedings{zhao2024estimating,
  title={Estimating Noisy Class Posterior with Part-level Labels for Noisy Label Learning},
  author={Zhao, Rui and Shi, Bin and Ruan, Jianfei and Pan, Tianze and Dong, Bo},
  booktitle={Proceedings of the IEEE/CVF Conference on Computer Vision and Pattern Recognition},
  pages={22809--22819},
  year={2024}
}

@inproceedings{humblot2024noisy,
  title={A noisy elephant in the room: Is your out-of-distribution detector robust to label noise?},
  author={Humblot-Renaux, Galadrielle and Escalera, Sergio and Moeslund, Thomas B},
  booktitle={Proceedings of the IEEE/CVF Conference on Computer Vision and Pattern Recognition},
  pages={22626--22636},
  year={2024}
}

@inproceedings{qiaounderstanding,
  title={Understanding Domain Generalization: A Noise Robustness Perspective},
  author={Qiao, Rui and Low, Bryan Kian Hsiang},
  year={2024},
  booktitle={The Twelfth International Conference on Learning Representations}
}

@inproceedings{ji2023drugood,
  title={Drugood: Out-of-distribution dataset curator and benchmark for ai-aided drug discovery--a focus on affinity prediction problems with noise annotations},
  author={Ji, Yuanfeng and Zhang, Lu and Wu, Jiaxiang and Wu, Bingzhe and Li, Lanqing and Huang, Long-Kai and Xu, Tingyang and Rong, Yu and Ren, Jie and Xue, Ding and others},
  booktitle={Proceedings of the AAAI Conference on Artificial Intelligence},
  volume={37},
  number={7},
  pages={8023--8031},
  year={2023}
}

@article{chang2023csot,
  title={Csot: Curriculum and structure-aware optimal transport for learning with noisy labels},
  author={Chang, Wanxing and Shi, Ye and Wang, Jingya},
  journal={Advances in Neural Information Processing Systems},
  volume={36},
  pages={8528--8541},
  year={2023}
}

@inproceedings{li2024feddiv,
  title={FedDiv: Collaborative Noise Filtering for Federated Learning with Noisy Labels},
  author={Li, Jichang and Li, Guanbin and Cheng, Hui and Liao, Zicheng and Yu, Yizhou},
  booktitle={Proceedings of the AAAI Conference on Artificial Intelligence},
  volume={38},
  number={4},
  pages={3118--3126},
  year={2024}
}

@inproceedings{he2016deep,
  title={Deep residual learning for image recognition},
  author={He, Kaiming and Zhang, Xiangyu and Ren, Shaoqing and Sun, Jian},
  booktitle={Proceedings of the IEEE conference on computer vision and pattern recognition},
  pages={770--778},
  year={2016}
}

@inproceedings{zheng2024exploiting,
  title={Exploiting Negative Samples: A Catalyst for Cohort Discovery in Healthcare Analytics},
  author={Zheng, Kaiping and Chua, Horng-Ruey and Herschel, Melanie and Jagadish, HV and Ooi, Beng Chin and Yip, James Wei Luen},
  booktitle={Forty-first International Conference on Machine Learning},
  year={2024}
}

@inproceedings{bucarelli2023leveraging,
  title={Leveraging inter-rater agreement for classification in the presence of noisy labels},
  author={Bucarelli, Maria Sofia and Cassano, Lucas and Siciliano, Federico and Mantrach, Amin and Silvestri, Fabrizio},
  booktitle={Proceedings of the IEEE/CVF Conference on Computer Vision and Pattern Recognition},
  pages={3439--3448},
  year={2023}
}

@inproceedings{yisource,
  title={When Source-Free Domain Adaptation Meets Learning with Noisy Labels},
  author={Yi, Li and Xu, Gezheng and Xu, Pengcheng and Li, Jiaqi and Pu, Ruizhi and Ling, Charles and McLeod, Ian and Wang, Boyu},
  year={2023},
  booktitle={The Eleventh International Conference on Learning Representations}
}

@article{liu2023ss,
  title={SS-Norm: Spectral-spatial normalization for single-domain generalization with application to retinal vessel segmentation},
  author={Liu, Yi-Peng and Zeng, Dongxu and Li, Zhanqing and Chen, Peng and Liang, Ronghua},
  journal={IET Image Processing},
  volume={17},
  number={7},
  pages={2168--2181},
  year={2023},
  publisher={Wiley Online Library}
}

@inproceedings{chensoftmatch,
  title={SoftMatch: Addressing the Quantity-Quality Tradeoff in Semi-supervised Learning},
  author={Chen, Hao and Tao, Ran and Fan, Yue and Wang, Yidong and Wang, Jindong and Schiele, Bernt and Xie, Xing and Raj, Bhiksha and Savvides, Marios},
  booktitle={The Eleventh International Conference on Learning Representations},
  year={2024}
}

@inproceedings{gulrajanisearch,
  title={In Search of Lost Domain Generalization},
  author={Gulrajani, Ishaan and Lopez-Paz, David},
  booktitle={International Conference on Learning Representations},
  year={2021}
}

@inproceedings{rodrigues2018deep,
  title={Deep learning from crowds},
  author={Rodrigues, Filipe and Pereira, Francisco},
  booktitle={Proceedings of the AAAI conference on artificial intelligence},
  volume={32},
  number={1},
  year={2018}
}

@inproceedings{DosovitskiyB0WZ21,
  title={An image is worth 16x16 words: Transformers for image recognition at scale},
  author={Alexey Dosovitskiy and
                  Lucas Beyer and
                  Alexander Kolesnikov and
                  Dirk Weissenborn and
                  Xiaohua Zhai and
                  Thomas Unterthiner and
                  Mostafa Dehghani and
                  Matthias Minderer and
                  Georg Heigold and
                  Sylvain Gelly and
                  Jakob Uszkoreit and
                  Neil Houlsby},
  booktitle={International Conference on Learning Representations},
  year={2021}
}

@inproceedings{teterwak2024ermimprovedbaselinedomain,
      title={ERM++: An Improved Baseline for Domain Generalization}, 
      author={Piotr Teterwak and Kuniaki Saito and Theodoros Tsiligkaridis and Kate Saenko and Bryan A. Plummer},
      year={2025},
      booktitle={Winter Conference on Applications of Computer Vision}
}

@article{arjovsky2019invariant,
  title={Invariant risk minimization},
  author={Arjovsky, Martin and Bottou, L{\'e}on and Gulrajani, Ishaan and Lopez-Paz, David},
  journal={arXiv preprint arXiv:1907.02893},
  year={2019}
}

@inproceedings{sagawadistributionally,
  title={Distributionally Robust Neural Networks},
  author={Sagawa, Shiori and Koh, Pang Wei and Hashimoto, Tatsunori B and Liang, Percy},
  booktitle={International Conference on Learning Representations},
  year={2019}
}

@inproceedings{krueger2021out,
  title={Out-of-distribution generalization via risk extrapolation (rex)},
  author={Krueger, David and Caballero, Ethan and Jacobsen, Joern-Henrik and Zhang, Amy and Binas, Jonathan and Zhang, Dinghuai and Le Priol, Remi and Courville, Aaron},
  booktitle={International conference on machine learning},
  pages={5815--5826},
  year={2021},
  organization={PMLR}
}

@inproceedings{fang2013unbiased,
  title={Unbiased metric learning: On the utilization of multiple datasets and web images for softening bias},
  author={Fang, Chen and Xu, Ye and Rockmore, Daniel N},
  booktitle={Proceedings of the IEEE International Conference on Computer Vision},
  pages={1657--1664},
  year={2013}
}

@inproceedings{venkateswara2017deep,
  title={Deep hashing network for unsupervised domain adaptation},
  author={Venkateswara, Hemanth and Eusebio, Jose and Chakraborty, Shayok and Panchanathan, Sethuraman},
  booktitle={Proceedings of the IEEE conference on computer vision and pattern recognition},
  pages={5018--5027},
  year={2017}
}

@article{jerant2000early,
  title={Early detection and treatment of skin cancer},
  author={Jerant, Anthony F and Johnson, Jennifer T and Sheridan, Catherine Demastes and Caffrey, Timothy J},
  journal={American family physician},
  volume={62},
  number={2},
  pages={357--368},
  year={2000}
}

@article{medmnistv2,
    title={MedMNIST v2-A large-scale lightweight benchmark for 2D and 3D biomedical image classification},
    author={Yang, Jiancheng and Shi, Rui and Wei, Donglai and Liu, Zequan and Zhao, Lin and Ke, Bilian and Pfister, Hanspeter and Ni, Bingbing},
    journal={Scientific Data},
    volume={10},
    number={1},
    pages={41},
    year={2023},
    publisher={Nature Publishing Group UK London}
}

@inproceedings{medmnistv1,
    title={MedMNIST Classification Decathlon: A Lightweight AutoML Benchmark for Medical Image Analysis},
    author={Yang, Jiancheng and Shi, Rui and Ni, Bingbing},
    booktitle={IEEE 18th International Symposium on Biomedical Imaging (ISBI)},
    pages={191--195},
    year={2021}
}

@article{bilic2023liver,
  title={The liver tumor segmentation benchmark (lits)},
  author={Bilic, Patrick and Christ, Patrick and Li, Hongwei Bran and Vorontsov, Eugene and Ben-Cohen, Avi and Kaissis, Georgios and Szeskin, Adi and Jacobs, Colin and Mamani, Gabriel Efrain Humpire and Chartrand, Gabriel and others},
  journal={Medical Image Analysis},
  volume={84},
  pages={102680},
  year={2023},
  publisher={Elsevier}
}

@article{ma2024sharpness,
  title={Sharpness-Aware Gradient Alignment for Domain Generalization with Noisy Labels in Intelligent Fault Diagnosis},
  author={Ma, Yulin and Yang, Jun and Yan, Ruqiang},
  journal={IEEE Transactions on Instrumentation and Measurement},
  year={2024},
  publisher={IEEE}
}

@article{sanyal2024accuracy,
  title={Accuracy on the wrong line: On the pitfalls of noisy data for out-of-distribution generalisation},
  author={Sanyal, Amartya and Hu, Yaxi and Yu, Yaodong and Ma, Yian and Wang, Yixin and Sch{\"o}lkopf, Bernhard},
  journal={arXiv preprint arXiv:2406.19049},
  year={2024}
}

@inproceedings{albert2022embedding,
  title={Embedding contrastive unsupervised features to cluster in-and out-of-distribution noise in corrupted image datasets},
  author={Albert, Paul and Arazo, Eric and O’Connor, Noel E and McGuinness, Kevin},
  booktitle={European Conference on Computer Vision},
  pages={402--419},
  year={2022},
  organization={Springer}
}

@inproceedings{albert2022addressing,
  title={Addressing out-of-distribution label noise in webly-labelled data},
  author={Albert, Paul and Ortego, Diego and Arazo, Eric and O'Connor, Noel E and McGuinness, Kevin},
  booktitle={Proceedings of the IEEE/CVF winter conference on applications of computer vision},
  pages={392--401},
  year={2022}
}

@article{graber2005diagnostic,
  title={Diagnostic error in internal medicine},
  author={Graber, Mark L and Franklin, Nancy and Gordon, Ruthanna},
  journal={Archives of internal medicine},
  volume={165},
  number={13},
  pages={1493--1499},
  year={2005},
  publisher={American Medical Association}
}

\end{document}